\documentclass{llncs}
\usepackage{graphicx}
\usepackage{booktabs}
\usepackage{epsfig}
\usepackage{graphicx}
\usepackage{amsmath}
\usepackage{amssymb}
\usepackage{textcomp}
\usepackage{enumitem} 
\usepackage{graphicx}
\usepackage{multirow}
\usepackage{array}
\usepackage{caption}
\usepackage[hyphens]{url}
\usepackage{balance}  
\usepackage{graphics} 
\usepackage{times}    
\usepackage{url}      
\usepackage{color}
\usepackage{verbatim}
\usepackage{pifont}
\usepackage{multirow}
\usepackage[lined,boxed,linesnumbered,ruled]{algorithm2e}
\usepackage{epstopdf}
\usepackage[]{algorithm2e}
\usepackage{xcolor,colortbl}
\usepackage{hyperref}

\usepackage{subfig}
\usepackage{cite}
\usepackage{inconsolata}

\usepackage{makecell}
\DontPrintSemicolon

\usepackage{etoolbox}
\usepackage{epstopdf}
\usepackage{graphicx}
\usepackage{color} 
\usepackage{verbatim}
\usepackage{amsmath}
\usepackage{multirow}
\usepackage{enumitem}
\usepackage{amssymb}
\usepackage[lined,boxed,linesnumbered,ruled]{algorithm2e}
\usepackage[]{algorithm2e}
\usepackage{epsfig}
\usepackage{commath} 
\usepackage{xcolor,colortbl}
\usepackage{caption}

\usepackage{url}
\usepackage{hyperref}
\usepackage{textcomp}
\usepackage{makecell}
\newtheorem{myremark}{Remark}
\newtheorem{myproof}{Proof}

\newtheorem{mydefinition}{Definition}

\newtheorem{mytheorem}{Theorem}

\usepackage[titletoc,title]{appendix}
\usepackage{mathtools,xparse}

\def \eg{{e.g.,}}
\def \ie{{i.e.,}}
\def \etc{{etc.}}

\def \one{{\it i)}}
\def \two{{\it ii)}}
\def \three{{\it iii)}}

\newcommand{\distance}[2]{||#2||_{#1}}
\newcommand{\classlabel}{cl}
\newcommand{\setofinputs}{{\tt T}}
\newcommand{\expectation}[2]{{\tt E}_{#1}[#2]}
\newcommand{\normball}{B}

\newcommand{\real}{\mathbb{R}}

\def \FastL0Attack{{\em FastL0Attack}}
\newcommand{\TRE}{\mathtt{L0\text{-}TRE}}

\newcommand{\commentout}[1]{}

\begin{document}
\title{Global Robustness Evaluation of Deep Neural Networks with Provable Guarantees for the $L_0$ Norm}


\author{
  Wenjie Ruan\inst{1} \and 
  Min Wu\inst{1} \and 
  Youcheng Sun\inst{1} \and
  Xiaowei Huang\inst{2} \\
  Daniel Kroening\inst{1} \and
  Marta Kwiatkowska\inst{1}
 }
  \institute{
  $^1$University of Oxford, UK \\
  \email{\{wenjie.ruan; min.wu; youcheng.sun\}@cs.ox.ac.uk}\\
  	\email{\{kroening; marta.kwiatkowska\}@cs.ox.ac.uk}\\
  $^2$University of Liverpool, UK\\
  \email{xiaowei.huang@liverpool.ac.uk} 
}

\maketitle         

\begin{abstract}
Deployment of deep neural networks (DNNs) in safety- or security-critical systems requires provable guarantees on their correct behaviour. A common requirement is robustness to adversarial 
perturbations in a neighbourhood around an input. In this paper we focus 
on the $L_0$ norm and aim to compute, for a trained DNN and an input, the 
maximal radius of a safe norm ball around the input within which there are no
adversarial examples. Then we define global robustness as an expectation 
of the maximal safe radius over a test data set. We first show that the 
problem is NP-hard, and then propose an approximate approach to iteratively 
compute lower and upper bounds on the network's robustness. 
The approach is \emph{anytime}, i.e., it returns intermediate bounds and 
robustness estimates that are gradually, but strictly, improved as the 
computation proceeds;  \emph{tensor-based}, i.e., the computation is 
conducted over a set of inputs simultaneously, instead of one by one, to 
enable efficient GPU computation; and has \emph{provable guarantees}, 
i.e., both the bounds and the robustness estimates can converge to their 
optimal values. Finally, we demonstrate the utility of the proposed 
approach in practice to compute tight bounds by applying and adapting 
the anytime algorithm to a set of challenging problems, including global 
robustness evaluation, competitive $L_0$ attacks, test case generation for DNNs, and local robustness evaluation on large-scale ImageNet DNNs.
We release the code of all case studies via
GitHub\footnote{The code is available in \url{https://github.com/TrustAI/L0-TRE}}.
\end{abstract}

\section{Introduction}

Deep neural networks (DNNs) have achieved significant breakthroughs in the past few years and are now being deployed in many applications. However, in safety-critical domains, where human lives are at stake, and security-critical applications, which often have significant financial risks, concerns have been raised about the reliability of this technique. 
In established industries, e.g., avionics and automotive, such concerns have to be addressed during the certification process before the deployment of the product. 
During the certification process, the manufacturer needs to demonstrate to the relevant certification authority, e.g., the European Aviation Safety Agency or the Vehicle Certification Agency, that the product behaves correctly with respect to a set of high-level requirements. For this purpose, it is necessary to develop techniques for discovering critical requirements and supporting the case that these requirements are met by the product.

Safety certification for DNNs is challenging owing to the black-box nature of DNNs and the lack of rigorous foundations. An important low-level requirement for DNNs is the robustness to input perturbations. DNNs have been shown to suffer from poor robustness because of their susceptibility to \emph{adversarial examples}~\cite{szegedy2014intriguing}. These are small modifications to an input, sometimes imperceptible to humans, that make the network unstable.
As a result, significant effort has been directed towards approaches for crafting adversarial examples or defending against them~\cite{goodfellow2014explaining, papernot2016limitations, carlini2017towards}. 
However, the cited approaches provide \emph{no} formal guarantees, i.e., no conclusion can be made whether adversarial examples remain or how close crafted adversarial examples are to the optimal ones.

Recent efforts in the area of automated verification~\cite{HKWW2017, katz2017reluplex} have instead focused on methods that generate adversarial examples, if they exist, and provide rigorous robustness proofs otherwise. 
These techniques rely on either a layer-by-layer exhaustive search of the neighbourhood of an image~\cite{HKWW2017}, or a reduction to a constraint solving problem by encoding the network as a set of constraints~\cite{katz2017reluplex}. Constraint-based approaches are limited to small networks. Exhaustive search, on the other hand, applies to large networks but suffers from the state-space explosion problem. To mitigate this, a Monte-Carlo tree search has been employed~\cite{WHK2017}. Moreover, a game-based approximate verification approach that can provide provable guarantees has been proposed\cite{wu2018game}.

This paper proposes a novel approach to quantify the robustness of DNNs that offers a balance between the guaranteed accuracy of the 
method (thus, a feature so far exclusive to formal approaches) and the efficiency of algorithms that search for adversarial examples (without providing any guarantees).
We consider the \emph{global} robustness problem, which is a generalisation of the local, pointwise robustness problem. Specifically, we define a maximum safe radius for every input and then evaluate robustness over a given test dataset, i.e., a finite set of inputs.  Global robustness is defined as the expected maximum safe radius over the test examples. We focus on the $L_0$ norm, which measures the distance between two matrices (e.g., two input images) by counting the number of elements (e.g., pixels) that are different. 

The key idea of our approach is to generate sequences of lower and upper bounds for 
global robustness. Our method is {\em anytime}, {\em tensor-based}, and offers {\em provable guarantees}. 
First, the method is \emph{anytime} in the sense that it can return intermediate results, including upper and lower bounds and robustness estimates. We prove that our approach can gradually, but strictly, improve these bounds and estimates as the computation proceeds. 
Second, it is \emph{tensor-based}. As we are working with a set of inputs, a straightforward approach is to perform robustness evaluation for the inputs individually and to then merge the results. However, this is inefficient, as the set of inputs is large. To exploit the parallelism offered by GPUs, our approach uses tensors. A tensor is a finite set of multi-dimensional arrays, and each element of the set represents one input. A good tensor-based algorithm uses tensor operations whenever possible.
Third, our approach offers \emph{provable guarantees}. We show that the intermediate bounds and the robustness estimates will converge to their optimal values in finite time, although this may be impractical for large networks. 


We implement our approach in a tool we name $\TRE$ (``{\em T}ensor-based {\em R}obustness {\em E}valuation for the $L_0$ Norm''), and conduct experiments on a set of challenging problems, including {\em Case Study 1:} global robustness evaluation; {\em Case Study 2:}~competitive $L_0$ attacks; {\em Case Study 3:}~test case generation; {\em Case Study 4:}~guidance for the design of robust DNN architectures; and {\em Case Study 5:}~saliency map generation for model interpretability and local robustness evaluation on five ImageNet DNNs including AlexNet, VGG-16/19 and ResNet-50/101.

All applications above require only simple adaptations of our method, e.g., slight modifications of the constraints or objective functions, or the addition of extra constraints. This demonstrates that our new technique is flexible enough to deliver a wide range of promising 
applications. The main contributions of this paper are as follows:
\begin{itemize}
	\item We propose a novel method to quantify global robustness of DNNs w.r.t.~the $L_0$-norm. This offers two key advantages, including \one~theoretical lower and upper bounds to guarantee its convergence; and \two~explicit tensor-based parallelisation on GPUs with high computational efficiency.
	
	\item With simple adaptations, we show the utility of the proposed method on a broad range of 
	applications, including \one~anytime global robustness evaluation; \two~competitive $L_0$ adversarial attacks; and \three~test case generation, etc.
	
	\item We perform a rigorous theoretical analysis and extensive empirical case studies to support the claims above. We test our tool on 15 different deep neural networks, including eight MNIST DNNs, two CIFAR-10 DNNs and five ImageNet DNNs.
	
\end{itemize}

\section{Problem Formulation}

Let $f: \real^n \rightarrow \real^m$ be an $N$-layer 
neural network such that, for a given input $x\in \real^n$, $f(x) = (c_1(x),c_2(x),\ldots,c_m(x))\in \real^m$ represents the confidence values for $m$ classification labels. Specifically, we have 
\begin{equation}
f(x) = f_N(f_{N-1}(\ldots f_1(x;W_1,b_1);W_2,b_2);\ldots );W_N,b_N)
\end{equation} 
where $W_i$ and $b_i$ for $i = 1,2,\ldots,N$ are learnable parameters, and $f_i(z_{i-1};W_{i-1},b_{i-1})$ is the function that maps the output of layer $i-1$, \ie~$z_{i-1}$, to the input of layer~$i$. Without loss of generality, we normalise the input to $x\in [0,1]^n$. The output $f(x)$ is usually normalised to be in $[0,1]^m$ with a softmax layer. We denote the classification label of input $x$ by $\classlabel(f,x) = \arg \max_{j = 1,\ldots,m} c_j(x)$. Note that both $f$ and $\classlabel$ can be generalised to work with a set $\setofinputs_0$ of inputs, i.e., $f(\setofinputs_0)$ and $\classlabel(f,\setofinputs_0)$, in the standard way.



\begin{mydefinition}[Safe Norm Ball]
	Given a  network $f: \real^n \rightarrow \real^m$, an input $x_0\in  \real^n$, a~distance metric $\distance{D}{\cdot}$ and a real number $d\in\real$, a \emph{norm ball} $\normball(f,x_0,\distance{D}{\cdot},d) $ is a subspace of $\real^n$ such that 
	\begin{equation}
	\normball(f,x_0,\distance{D}{\cdot},d) =\{x ~|~ \distance{D}{x_0-x} \leq d\}.
	\end{equation}
	%
	%
	The number $d$ is called the radius of $\normball(f,x_0,\distance{D}{\cdot},d)$. A norm ball $\normball(f,x_0,\distance{D}{\cdot},d)$ is \emph{safe} if for all $ x\in \normball(f,x_0,\distance{D}{\cdot},d)$ we have $\classlabel(f,x) = \classlabel(f,x_0)$. 
\end{mydefinition}

Intuitively, a norm ball $\normball(f,x_0,\distance{D}{\cdot},d) $ includes all inputs whose distance to $x_0$, measured by a metric $\distance{D}{\cdot}$, is within $d$.

\begin{mydefinition}[Maximum Radius of a Safe Norm Ball]
	Let $d$ be the radius of a safe norm ball $\normball(f,x_0,\distance{D}{\cdot},d)$. If for all $d' > d$ we have that  $\normball(f,x_0,\distance{D}{\cdot},d')$ is not safe, then $d$ is called the \emph{maximum safe radius}, denoted by $d_m(f,x_0,\distance{D}{\cdot})$. Formally, 
	\begin{equation}
	d_m(f,x_0,\distance{D}{\cdot}) = \arg\sup_{d} \{ \normball(f,x_0,\distance{D}{\cdot},d) \text{ is safe} ~|~ d \in \real, d>0  \} .
	\end{equation} 
	%
\end{mydefinition}

We 
define the (\emph{global}) robustness evaluation problem over a testing dataset $\setofinputs$, which is a set of i.i.d.~inputs sampled from a distribution $\mu$ representing the problem $f$ is working on. We use $|\setofinputs|$ to denote the number of inputs in $\setofinputs$. When $|\setofinputs| =1$, we 
call it \emph{local} robustness. 

\begin{mydefinition}[Robustness Evaluation]\label{def:localrobustness}
	Given a network $f$, a finite set $\setofinputs_0$ of inputs, and a distance metric $\distance{D}{\cdot}$, the \emph{robustness evaluation}, denoted as $R(f,\setofinputs_0,\distance{D}{\cdot})$, is an optimisation problem: 
	\vspace{-1ex}
	\begin{equation}
	\label{eq:robustness}
	\begin{split}
	&\displaystyle \min_{\setofinputs} \distance{D}{\setofinputs_0 - \setofinputs} \\
	\text{ s.t. }& ~~
	~~\classlabel(f,x_i) \neq \classlabel(f,x_{0,i}) \quad\text{for~} i=1, \ldots, |\setofinputs_0|
	\end{split}
	\end{equation}
	where  $ \setofinputs = (x_i)_{i=1 \ldots |\setofinputs_0|}$, $\setofinputs_0 = (x_{0,i})_{i=1 \ldots |\setofinputs_0|} $, and $x_i,x_{0,i}\in [0,1]^{n}$. 
	
\end{mydefinition}	
Intuitively, we aim to find a minimum distance between the original set $\setofinputs_0$ and a new, homogeneous set $\setofinputs$ of inputs such that all inputs in $\setofinputs_0$ are misclassified. The two sets $\setofinputs_0$ and $\setofinputs$ are homogeneous if they have the same number of elements and their corresponding elements are of the same type.


\paragraph{$L_0$  Norm}\label{sec:l0distance}

The distance metric $\distance{D}{\cdot}$ can be any mapping $\distance{D}{\cdot}: \mathbb{R}^n \times \mathbb{R}^n \rightarrow [0, \infty]$ that satisfies the metric conditions. In this paper, we focus on the $L_0$ metric. For two inputs $x_0$ and $x$, their $L_0$ distance, denoted as $\distance{0}{x-x_0}$, is the number of elements that are different. When working with test datasets, we define 
\begin{equation}
\begin{array}{lcll}
\distance{0}{\setofinputs-\setofinputs_0} & = & \expectation{x_0\in \setofinputs_0}{\distance{0}{x-x_0}} & \text{(our definition)}\\
& = & \frac{1}{|\setofinputs_0|}\sum_{x_0\in \setofinputs_0}\distance{0}{x-x_0} & \text{(all inputs in $\setofinputs_0$ are i.i.d.)}
\end{array}
\end{equation}
where $x \in \setofinputs$ is a homogeneous input of $x_0 \in \setofinputs_0$.
While other norms such as $L_1$, $L_2$ and $L_\infty$ have been widely applied for generating adversarial examples~\cite{papernot2016limitations,kurakin2016adversarial}, studies based on the $L_0$ norm 
are few and far between. In the Appendix, we justify why $L_0$ is the appropriate metric for our goals.

\section{Anytime Robustness Evaluation}\label{sec:anytime}


The accurate evaluation of robustness in Definition~\ref{def:localrobustness} is hard in terms of $L_0$-norm distance. In Appendix~\ref{sec:np}, we give the computational complexity and prove its NP-hardness.

In this paper, we propose to compute lower and upper bounds, and then gradually, but \emph{strictly}, improve the bounds so that the gap between them can eventually be closed in finite time. Although the realistic running time can be long, this \emph{anytime} approach provides pragmatic means to track progress.
%
Experimental results in Section~\ref{sec:experiments} show that our approach is able to achieve \emph{tight} bounds \emph{efficiently} in practice. 

\begin{mydefinition}[Sequences of Bounds]\label{sec:boundsequence}
	Given a robustness evaluation problem $R(f,\setofinputs_0,\allowbreak\mbox{$\distance{D}{\cdot}$})$, a sequence $\mathcal{L}(\setofinputs_0) = \{l_1, l_2,\ldots,l_k\}\in\real$ is an incremental lower bound sequence if, for all $ 1\leq i < j\leq k$, we have $l_i\leq l_{j}\leq R(f,\setofinputs_0,\distance{D}{\cdot})$. The sequence is strict, denoted as $\mathcal{L}_s(\setofinputs_0)$, if for all $1\leq i < j\leq k$, we have either $l_i< l_{j}$ or $l_i=l_j = R(f,\setofinputs_0,\distance{D}{\cdot})$. 
	%
	Similarly, we can define a decremental upper bound sequence $\mathcal{U}(\setofinputs_0)$ and a strict decremental upper bound sequence $\mathcal{U}_s(\setofinputs_0)$. 
\end{mydefinition}

We will, in Section~\ref{sec:solution}, introduce our algorithms on computing these two sequences of lower and upper bounds. For now, assume they exist, then at a certain time $t > 0$, 
$l_t \leq R(f,\setofinputs_0,\distance{D}{\cdot}) \leq u_t $
holds.

\begin{mydefinition}[Anytime Robustness Evaluation]
	For a given range $[l_t,u_t]$, 
	we define its centre and radius as follows. 
	\begin{equation}
	U_c(l_t, u_t) = \dfrac{1}{2}(l_t+u_t) \quad\text{and}\quad U_r(l_t,u_t) = \dfrac{1}{2} (u_t-l_t).
	\end{equation}
	The anytime evaluation of $R(f,\setofinputs_0,\distance{D}{\cdot})$ at time $t$, denoted as $R_t(f,\setofinputs_0,\distance{D}{\cdot})$, is the pair $(U_c(l_t, u_t),U_r(l_t,u_t))$. 
\end{mydefinition}
The anytime evaluation will be returned whenever the computational procedure is interrupted. Intuitively, we use 
$U_c(l_t, u_t)$ to represent the current estimation, and 
$U_r(l_t,u_t))$ to represent its error bound. Essentially, we can bound the true robustness $R(f,\setofinputs_0,\allowbreak\mbox{$\distance{D}{\cdot}$})$ via the anytime robustness evaluation.
Let $f$ be a network and $\distance{D}{\cdot}$ a distance metric. 
	At any time $t>0$, the anytime evaluation $R_t(f,\setofinputs_0,\distance{D}{\cdot})=(U_c(l_t, u_t),\allowbreak U_r(l_t,u_t))$ such that 	
	\begin{equation}
	\begin{array}{c}
	U_c(l_t, u_t) - U_r(l_t,u_t) \leq R(f,\setofinputs_0,\distance{D}{\cdot}) \leq U_c(l_t, u_t) + U_r(l_t,u_t).
	\end{array}
	\end{equation}

\section{Tensor-based Algorithms for Upper and Lower Bounds}
\label{sec:solution}

We present our approach to generate the sequences of bounds.

\begin{mydefinition}[Complete Set of Subspaces for an Input] \label{def:complete}
	Given an input $x_0\in [0,1]^n$ and a set of $t$ dimensions $T\subseteq \{1,...,n\}$ such that $|T| = t$, the subspace for $x_0$, denoted by $X_{x_0,T}$, is a set of inputs $x\in [0,1]^n$ 
	such that $x(i)\in [0,1]$ for $i\in T$ and $x(i)=x_0(i)$ for $i\in \{1,...,n\}\setminus T$. 
	Furthermore, given an input $x_0\in [0,1]^n$ and a number $t\leq n$, we define 
	\begin{equation}
	\mathcal{X}(x_0,t) = \{X_{x_0,T} ~|~ T\subseteq \{1,...,n\}, |T| = t \}
	\end{equation}
	as the complete set of subspaces for input $x_0$.


\end{mydefinition} 
Intuitively, elements in $X_{x_0,T}$ share the same value with $x_0$ on the dimensions other than $T$, and may take any legal value for the dimensions in $T$. Moreover, $\mathcal{X}(x_0,t)$ includes all  sets $X_{x_0,T}$ for any possible combination $T$ with $t$ dimensions.

Next, we define the subspace sensitivity for a subspace w.r.t. a network $f$, an input~$x_0$ and a test dataset $\setofinputs_0$. Recall that $f(x) = (c_1(x),c_2(x),\ldots,c_m(x))$. 

\begin{mydefinition}[Subspace Sensitivity] \label{def:sensitivity}
	Given an input subspace $X\subseteq [0,1]^n$, an input $x_0\in [0,1]^n$ and a label $j$, the subspace sensitivity w.r.t. $X$, $x_0$, and $j$ is defined as 
	\begin{equation}\label{equ:subspacesensitivity}
	S(X,x_0,j) =c_j(x_0) - \inf_{x\in X} c_j(x).
	\end{equation}
	Let $t$ be an integer.
	We define the subspace sensitivity for $\setofinputs_0$ and $t$ as
	\begin{equation} \label{eqn:9}
	S(\setofinputs_0,t) = (S(X_{x_0},x_0,j_{x_0}))_{X_{x_0}\in \mathcal{X}(x_0,t), x_0\in \setofinputs_0}
	\end{equation} 
	where $j_{x_0}=\arg\max_{i\in\{1,...,m\}}c_i(x_0)$ is the classification label of $x_0$ by network $f$. 
\end{mydefinition} 
Intuitively, $S(X,x_0,j)$ is the maximal decrease of confidence value of the output label~$j$ that can be witnessed from the set $X$, and $S(\setofinputs_0,t)$ is the two-dimensional array of the maximal decreases of confidence values of the classification labels for all subspaces in $\mathcal{X}(x_0,t)$ and all inputs in $\setofinputs_0$. 
It is not hard to see that $S(X,x_0,j)\geq 0$.

Given a test dataset $\setofinputs_0$ and an integer $t > 0$, the number of elements in $S(\setofinputs_0,t)$ is in $O(|\setofinputs_0|\cdot n^{t})$, i.e., polynomial in $|\setofinputs_0|$ and exponential in~$t$. Note that, by Equation~(\ref{equ:subspacesensitivity}), every element in $S(\setofinputs_0,t)$ represents an optimisation problem. E.g., for $\setofinputs_0$, a set of 20 MNIST images, and $t=1$, this would be $28\times 28\times 20 = 15,\!680$ one-dimensional optimisation problems. In the next section, we give a tensor-based formulation and an algorithm to solve this challenging problem via GPU parallelisation.


\subsection{Tensor-based Parallelisation for Computing Subspace Sensitivity}
\label{sec:5.1}

A tensor $\mathcal{T}\in \mathbb{R}^{I_1\times I_2 \times \ldots \times I_N}$ in an $N$-dimensional space is a mathematical object that has $\prod_{m = 1}^{N} I_m $ components and obeys certain transformation rules. Intuitively, tensors are generalisations of vectors (\ie~one index) and matrices (\ie~two indices) to an arbitrary number of indices. Many state-of-the-art deep learning libraries, such as Tensorflow and Keras, are utilising the tensor format to parallelise the computation with GPUs. However, it is nontrivial to write an algorithm working with tensors due to the limited set of operations on tensors.

The basic idea of our algorithm is to transform a set of nonlinear, noncovex optimisation problems as given in Equation~(\ref{eqn:9})  into a tensor formulation, and  solve a set of optimisation problems via a few DNN queries. First, we introduce the following operations on tensors we use in our algorithm.

\begin{mydefinition}[Mode-$n$ Unfolding and Folding]
	Given a tensor $\mathcal{T}\in \mathbb{R}^{I_1 \times I_2 \times \ldots \times I_N}$, the mode-n unfolding of $\mathcal{T}$ is a matrix $\mathbf{U}_{[n]}(\mathcal{T}) \in \mathbb{R}^{I_n\times I_M}$ such that $M = \prod_{\substack{k=1,k \neq n}}^N I_k$ and $\mathbf{U}_{[n]}(\mathcal{T})$ is defined by the mapping from element $(i_1, i_2, \ldots, i_N)$ to $(i_n, j)$, with
	$$
	j = \sum_{k=1,\\k\neq n}^{N}i_k \times \prod_{m = k+1,k\neq n}^{N} I_m .
	$$
	Accordingly, the tensor folding $\mathbf{F}$ folds an unfolded tensor back from a matrix to a full tensor. Tensor unfolding and folding are dual operations and link tensors and matrices.
\end{mydefinition}

Given a neural network $f$, a number $t$ and a test dataset $ \setofinputs_0$, each $x_i \in \setofinputs_0$ generates a complete set $\mathcal{X}(x_i,t)$ of subspaces.
Let $|\setofinputs_0| = p$ and $|\mathcal{X}(x_i,t)| = k$. Note that for different $x_i$ and $x_j$, we have $|\mathcal{X}(x_i,t)|=|\mathcal{X}(x_j,t)|$. 
Given an error tolerance $\epsilon>0$, by applying grid search, we can recursively sample $\Delta = 1/\epsilon$  numbers in each dimension, and turn each subspace $X_{x_i}\in \mathcal{X}(x_i,t)$ into a two-dimensional grid $\mathcal{G}(X_{x_i}) \in \mathbb{R}^{n\times \Delta^t}$. We can formulate the following 
tensor:
\begin{equation}
\mathcal{T}(\setofinputs_0,t) = \text{Tensor}((\mathcal{G}(X_{x_i}))_{x_i\in \setofinputs_0, X_{x_i} \in \mathcal{X}(x_i,t)}) \in \mathbb{R}^{n\times \Delta^t\times p\times k}
\end{equation}

In Sec.~\ref{sec:CA}, we show that grid search provides the guarantee of reaching the global minimum by utilizing the Lipschitz continuity in DNNs. 


Then, we apply the mode-1 tensor unfolding operation to have $\mathbf{T}_{[1]}(\mathcal{T}(\setofinputs_0,t)) \in \mathbb{R}^{n\times M}$ such that $M = \Delta^t \cdot p \cdot k$. Then this tensor can be fed into the DNN $f$ to obtain
\begin{equation}\label{equ:DNNquery1}
Y(\setofinputs_0,t) = f(\mathbf{T}_{[1]}(\mathcal{T}(\setofinputs_0,t))) \in \mathbb{R}^M .
\end{equation}

After computing $Y(\setofinputs_0,t)$, we apply a tensor folding operation to obtain
\begin{equation}
\mathcal{Y}(\setofinputs_0,t) = \mathbf{F}(Y(\setofinputs_0,t)) \in \mathbb{R}^{\Delta^t \times p \times k}.  
\end{equation}
Here, we should note the difference between $\mathbb{R}^{\Delta^t \cdot p \cdot k}$ and $\mathbb{R}^{\Delta^t \times p \times k}$, with the former being a one-dimensional array and the latter a  tensor. 
On $\mathcal{Y}(\setofinputs_0,t)$, we search the minimum values along the first dimension to obtain\footnote{Here we use a Matlab notation $\min(Y,k)$, which computes the minimum values over the $k$-th dimension for a multi-dimensional array $Y$. Other notation to appear later is similar.} 
\begin{equation}
V(\setofinputs_0,t)_\mathrm{min} = \min (\mathcal{Y}(\setofinputs_0,t),1) \in \mathbb{R}^{p\times k}
\end{equation} 
Thus, we have now solved all $p\times k$ optimisation problems. We then construct the tensor 
\vspace*{-0.75ex}
\begin{equation}
V(\setofinputs_0,t) = (\overbrace{c_{j_{x_i}}(x_i),...,c_{j_{x_i}}(x_i)}^k)_{x_i\in \setofinputs_0} \in \mathbb{R}^{p\times k}
\end{equation}
from the set $\setofinputs_0$. Recall that $j_{x_i}=\arg\max_{k\in\{1,...,m\}}c_k(x_i)$. 
Intuitively, $V(\setofinputs_0,t)$ is the tensor that contains the starting points of the optimisation problems and $V(\setofinputs_0,t)_\mathrm{min}$ the resulting optimal values. 
%
%
%
The following theorem shows the correctness of our computation, where $S(\setofinputs_0,t)$ has been defined in Definition~\ref{def:sensitivity}. 

\begin{mytheorem}
	Let $\setofinputs_0$ be a test dataset and $t$ an integer. We have  $S(\setofinputs_0,t) = V(\setofinputs_0,t) - V(\setofinputs_0,t)_\mathrm{min}$.

	
\end{mytheorem}
To perform the computation above, we only need a single DNN query in Equation~(\ref{equ:DNNquery1}).



\subsection{Tensor-based Parallelisation for Computing Lower and Upper Bounds}
\label{sec:bounds}


Let 
${\cal S}(\setofinputs_0,t)\in \mathbb{R}^{n\times p\times k}$ be the tensor obtained by replacing every element in $S(\setofinputs_0,t)$ with their corresponding inputs that, according to the computation of $V(\setofinputs_0,t)_\mathrm{min}$, cause the largest decreases on the confidence values of the classification labels. We call 
${\cal S}(\setofinputs_0,t)$ the \emph{solution tensor} of 
$S(\setofinputs_0,t)$.
The computation of ${\cal S}(\setofinputs_0,t)$
can 
be done using very few tensor operations over $\mathcal{T}(\setofinputs_0,t)$ and $\mathcal{Y}(\setofinputs_0,t) $, which have been given in Section~\ref{sec:5.1}. We~omit the details. 




\paragraph{Lower Bounds}

We reorder $S(\setofinputs_0,t)$ and ${\cal S}(\setofinputs_0,t)$ w.r.t. the decreased values in $S(\setofinputs_0,t)$. 
Then, we retrieve the first row of the third dimension in tensor ${\cal S}(\setofinputs_0,t)$, \ie~${\cal S}(\setofinputs_0,t)[:,\allowbreak :,\allowbreak 1] \in \mathbb{R}^{n\times p}$, and check whether $\classlabel(f,{\cal S}(\setofinputs_0,t)[:,:,1] ) = \classlabel(f,\setofinputs_0)$. The result is an array of Boolean values, each of which is associated with an input $x_i\in \setofinputs_0$. 
If any element associated with $x_i$ in the resulting array is $\mathit{false}$, we conclude that $d_m(f,x_i,\distance{D}{\cdot}) = t-1$, i.e., the maximum safe radius has been obtained and the computation for $x_i$ has converged.
On the other hand, if the element associated with $x_i$ is $\mathit{true}$, we update the lower bound for $x_i$ to $t$. 
After computing ${\cal S}(\setofinputs_0,t)$, no further DNN query is needed to compute the lower bounds.

\paragraph{Upper Bounds}


The upper bounds are computed by iteratively applying perturbations based on the matrix ${\cal S}(\setofinputs_0,t)$ for every input in  $\setofinputs_0$ until a misclassification occurs. However, doing this sequentially for all inputs would be inefficient, since we need to query the network $f$ after every perturbation on each image.

We present an efficient tensor-based algorithm, which enables GPU parallelisation. The key idea is to construct a new tensor $\mathcal{N} \in \mathbb{R}^{n\times p \times k}$ to maintain all the accumulated perturbations over the original inputs $\setofinputs$. 
\begin{itemize}
	\item Initialisation: $\mathcal{N}[:,:,1] ={\cal S}(\setofinputs_0,t)[:,:,1]$.
	\item Iteratively construct the $i$-th row until $i=k$: 
	\begin{equation}
	\begin{array}{lcl}
	\mathcal{N}[:,:,i] &= &\{\mathcal{N}[:,:,i-1] \boxminus \{\mathcal{N}[:,:,i-1] \Cap {\cal S}(\setofinputs_0,t)[:,:,i] \}\}  \\
	&& \Cup \{{\cal S}(\setofinputs_0,t)[:,:,i] \boxminus \{\mathcal{N}[:,:,i-1] \Cap{\cal S}(\setofinputs_0,t)[:,:,i] \}\} 
	\end{array}
	\end{equation}
\end{itemize}
where $\boxminus$, $\Cap$, and $\Cup$ are tensor operations: $\mathcal{N}_1\boxminus \mathcal{N}_2$ removes the corresponding non-zero elements in $\mathcal{N}_2$ from $\mathcal{N}_1$; further, $\mathcal{N}_1\Cap \mathcal{N}_2$ retains those elements that have the same values and sets the other elements to 0; finally, $\mathcal{N}_1\Cup \mathcal{N}_2$ merges the non-zero elements from two tensors.  The two operands of these operations are required to have the same type.
Intuitively, $\mathcal{N}[:,:,i]$ represents the result of applying the first $i$ perturbations recorded in ${\cal S}(\setofinputs_0,t)[:,:,1:i]$.

Subsequently, we unfold $\mathcal{N}$ and pass the result to the DNN $f$, which yields the classification labels $Y(\mathbf{U}_{[1]}(\mathcal{N})) \in \{1,\ldots,m\}^{p \cdot k}$. After that, a tensor folding operation is applied to obtain ${\cal Y}(\mathbf{U}_{[1]}(\mathcal{N})) \in  \{1,\ldots,m\}^{p\times k}$. 
%
Finally, we can compute the minimum column index along each row such that misclassification happens, denoted by $\{m_1,m_2,...,m_p\}$ such that $1\leq m_i \leq k$. Then we let
\begin{equation}
\setofinputs = \{\mathcal{N}_{:,i,m_i} \in \mathbb{R}^{n\times p}~|~  x_i \in \setofinputs_0\}\,,
\end{equation}
which is the optimal set of inputs as required in Definition~\ref{def:localrobustness}. 

%
After computing ${\cal S}(\setofinputs_0,t)$, we only need one further DNN query to obtain all upper bounds for a given test dataset $\setofinputs_0$.


\newcommand{\setofgaps}{{\tt L}}

\paragraph{Tightening the Upper Bounds}

There may be redundancies in $\setofinputs-\setofinputs_0$, i.e., not all the changes in $\setofinputs-\setofinputs_0$ are necessary to observe misclassification. We therefore reduce the redundancies and thereby tighten the upper bounds. We reduce the tightening problem to an optimisation problem similar to that of Definition~\ref{def:localrobustness}, which enables us to reuse the tensor-based algorithms given above.

Assume that $x_{0,i}$ and $x_i$ are two corresponding inputs in $\setofinputs$ and $\setofinputs_0$, respectively, for $i\in \{1,\ldots,|\setofinputs_0|\}$. By abuse of notation, we let $z_{0,i} = x_{0,i} - x_i $ be the part of $x_{0,i}$ on which  $x_{0,i}$ and $x_i$  are different, and $l_{0,i} = x_i \Cap x_{0,i}$ be the part of $x_{0,i}$ on which  $x_{0,1}$ and $x_i$   are the same.  Therefore, $x_{0,i} = z_{0,i}  \Cup  l_{0,i}$. 


\begin{mydefinition}[Tightening the Upper Bounds] \label{def:tightenUpperBounds}
	Given a network $f$, a finite test dataset $\setofinputs_0$  with their upper bounds $\setofinputs$, and a distance metric $\distance{D}{\cdot}$, the tightening problem is an optimisation problem: 
	\begin{equation}\label{equ:tighten}
	\begin{split}
	&\displaystyle \min_{\setofgaps_1  } \distance{D}{\setofgaps_0 - \setofgaps_1} \\
	\text{ s.t. }& ~~
	\classlabel(f,z_{0,i} \Cup l_{1,i}) \neq \classlabel(f,z_{0,i}\Cup l_{0,i}) \quad\text{for } i=1, \ldots, |\setofgaps_0|
	\end{split}
	\end{equation}
	where  $ \setofgaps_0 = (l_{0,i})_{i=1 \ldots |\setofinputs_0|}$, $\setofgaps_1 = (l_{1,i})_{i=1 \ldots |\setofgaps_0|} $, $l_{1,i}, l_{0,i} \in [0,1]^{|\setofgaps_0|}$ have the same shape. 
\end{mydefinition}	
To solve this optimisation problem, we can re-use the tensor-based algorithm for computing lower bounds with minor modifications to the DNN query: Before querying DNN, we apply the $\Cup$ operation to merge with $z_{0,i}$, as suggested by Equation~(\ref{equ:tighten}).  

\subsection{Convergence}
\label{sec:CA}

We perform convergence analysis of the proposed method. For simplicity,  in the proofs we consider the case of a single input~$x_0$. The convergence guarantee can be extended easily to a finite set.
%
%
We first show that grid search can guarantee to find the global optimum given a certain error bound based on the assumption that the neural network satisfies the Lipschitz condition as proved in~\cite{WHK2017,PRGS2017}.

\begin{mytheorem}[Guarantee of the global minimum of grid search]\label{proof-0}
	Assume a neural network $f(x): {[0,1]}^n \to \mathbb{R}^m$ is Lipschitz continuous w.r.t. a norm metric $||\cdot||_D$ and its Lipschitz constant is $K$. By 
	recursively sampling $\Delta = 1/\epsilon$ in each dimension, denoted as $\mathcal{X} = \{x_1,...,x_{\Delta^n}\}$, the following relation holds: 
	$$
	||f_\mathrm{opt}(x^*) - \min_{x\in\mathcal{X}}f(x)||_D \,\leq\, K\cdot||\dfrac{\epsilon}{2}\mathbf{I}_n||_D
	$$
	where $f_\mathrm{opt}(x^*)$ represents the global minimum value, $\min_{x\in\mathcal{X}}f(x)$ denotes the minimum value returned by grid search, and $\mathbf{I}_n \in \mathbb{R}^{n\times n}$ is an all-ones matrix.
\end{mytheorem}

\begin{myproof}
	Based on the Lipschitz continuity assumption of $f$, we have
	$$
	||f(x_1) - f(x_2)||_D \,\leq\, K\cdot||x_1 - x_2||_D
	$$
	The $\epsilon$ grid search guarantees $\forall \tilde{x}\in [0,1]^n, \exists x \in \mathcal{X}$ such that $||x^*-x||_D \leq ||\dfrac{\epsilon}{2}\mathbf{I}_n||_D$, denoted as $\mathcal{X}(\tilde{x})$. Thus the theorem holds as we can always find  $\mathcal{X}(x^*)$ from the sampled set $\mathcal{X}$ for the global minimum $x^*$.
\end{myproof}

As shown in Sec.~\ref{sec:bounds}, in each iteration, we apply the grid search to verify the safety of the DNNs (meaning that we preclude adversarial examples) given a lower bound. In combination with Theorem~\ref{proof-0}, we arrive at the following, which shows the safety guarantee for the lower bounds. 

\begin{mytheorem}[Guarantee for Lower Bounds] \label{thm:lowerboundproof}
	Let $f$ denote a DNN and let $x_0\in [0,1]^n$ be an input. If our method generates a lower bound $l(f,x_0)$, then $\classlabel(f,x) =\classlabel(f,x_0)$ for all $x$ such that $||x-x_0||_0\leq l(f,x_0)$. I.e., $f$~is guaranteed to be safe for any pixel perturbations with at most $l(f,x_0)$ pixels.
\end{mytheorem}


Theorem~\ref{thm:lowerboundproof} (proof in Appendix~\ref{sec:proof-1}) shows that the lower bounds generated by our algorithm are the lower bounds of $d_m(f,x_0,\distance{D}{\cdot})$. We gradually increase $t = l(f,x_0)$ and re-run the lower bound generation algorithm. Because the number of dimensions of an input is finite, the distance to an adversarial example is also finite. Therefore, the lower bound generation algorithm converges eventually.

\begin{mytheorem}[Guarantee for Upper Bounds] \label{thm:upp}
	Let $f$ denote a DNN and $x_0\in [0,1]^n$ denote an input. Let $u_{i}(f,x_0)$ be an upper bound generated by our algorithm for any $i > 0$. Then we have $u_{i+1}(f,x_0) \leq u_i(f,x_0)$ for all $i > 0$, and  $\lim_{i\mapsto \infty} u_i(f,x_0) = d_m(f,x_0,\distance{D}{\cdot})$. 
\end{mytheorem}


The three key ingredients to show that the upper bounds decrease monotonically are: \one~the complete subspaces generated at $t = i$ are always included in the complete subspaces at $t = i+1$; \two~the pixel perturbation from a subspace with higher priority always results in a larger confidence decrease than those with lower priority; and \three~the tightening strategy is able to exclude the redundant pixel perturbations. The details of the proof for Theorem~\ref{thm:upp} are in Appendix~\ref{sec:proof-2}. Finally, we can show that the radius of $[l_i,u_i]$ will converge to 0 deterministically (see Appendix~\ref{sec:proof-3}).








\section{Experimental Results}\label{sec:experiments}


We report experimental evidence for the utility of our algorithm. Some experiments require simple modifications of the optimisation problem given in Definition~\ref{def:localrobustness}, e.g., small changes to the constraints. No significant modification to our algorithm is needed process to these variants. In this section, we use five case studies to demonstrate the broad applicability of our tool~\footnote{The Case Study Four and Case Study Five are available in the Appendix}.

\subsection{Case Study One: Convergence Analysis and  Global Robustness Evaluation}


\begin{figure}[t]
    \hspace*{-0.3cm}
	\begin{minipage}{0.37\linewidth}
		\centering
		\includegraphics[width=1\linewidth]{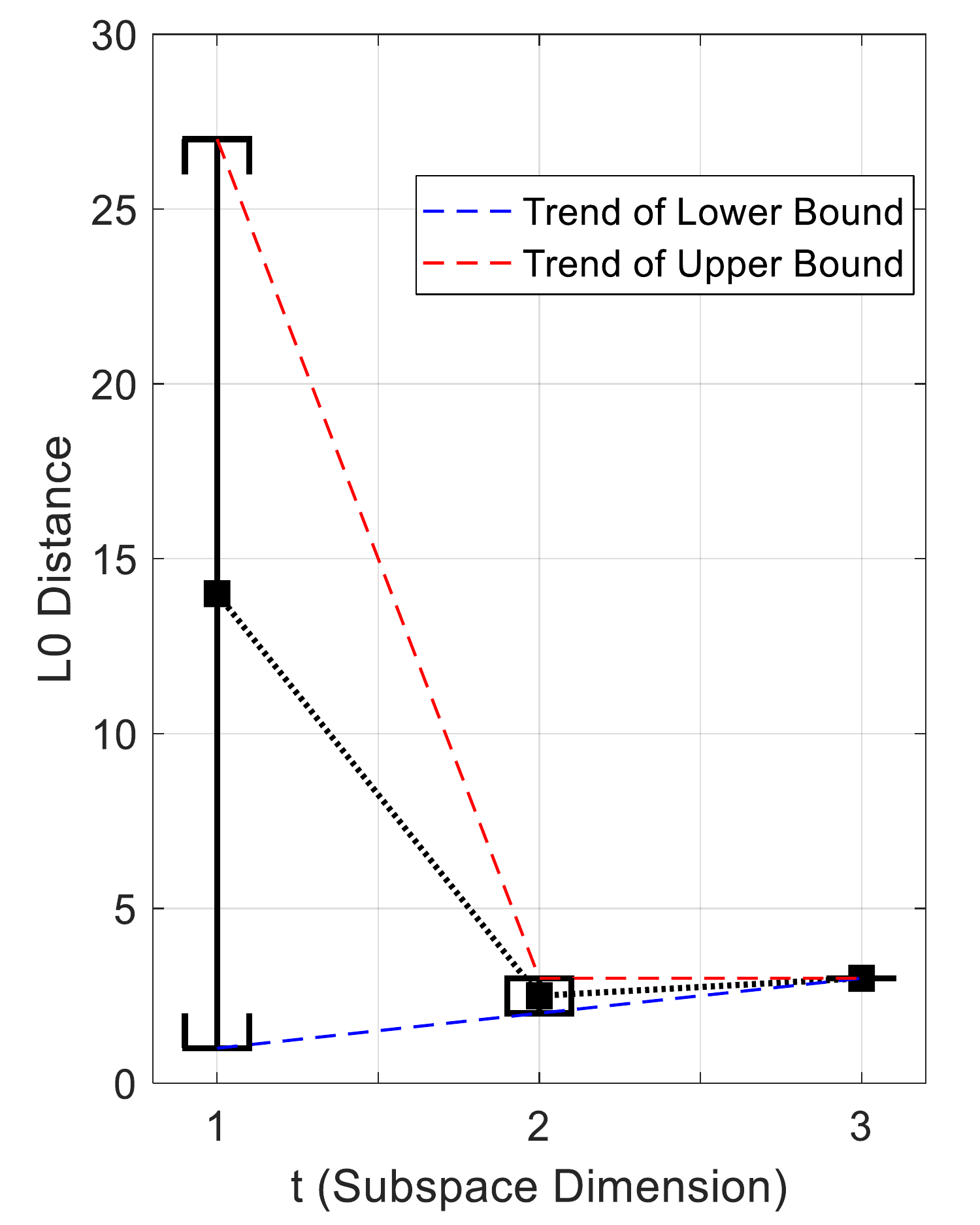}
		\text{(a)}
	\end{minipage}
	\begin{minipage}{0.37\linewidth}
		\centering
		\includegraphics[width=1\linewidth]{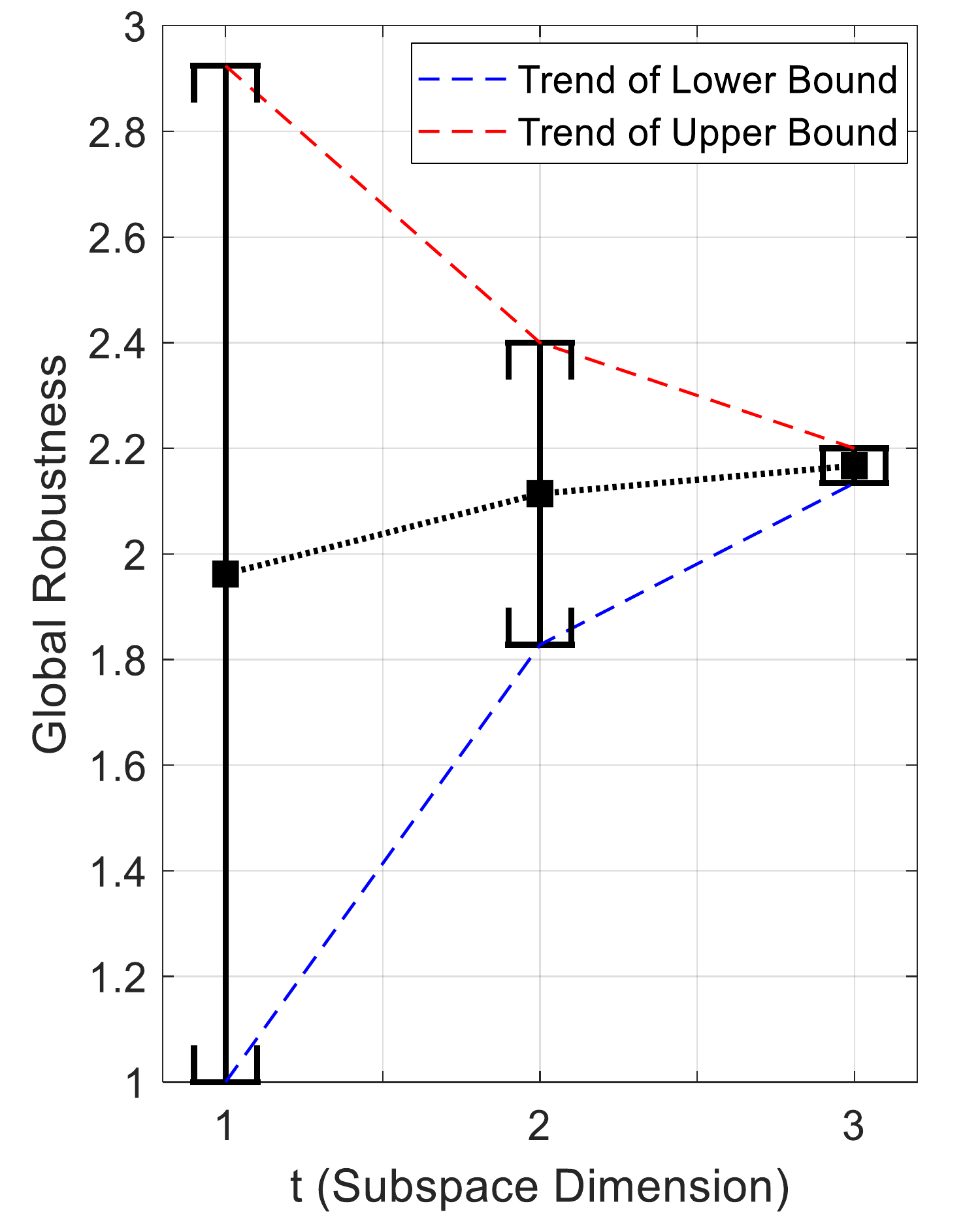}
		\text{(b)}
	\end{minipage}
	\begin{minipage}{0.24\linewidth}
		\centering
		\includegraphics[width=1\linewidth]{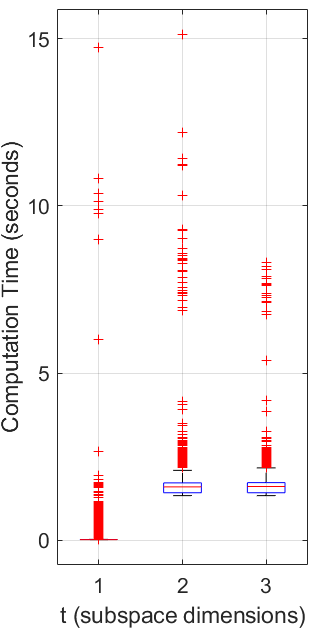}
		\text{(c)}
	\end{minipage}
		\vspace{-1mm}
	\caption{(a) Convergence of lower bound, upper bound, and estimation of $d_m$ for one image; (b)~Convergence of lower bound, upper bound, and estimation of global robustness; (c)~Boxplots of the computational time for $t \in \{ 1, 2, 3 \}$}
	\label{fig-1}
	\vspace{-2mm}
\end{figure}

\begin{figure}[t]
	\begin{minipage}{0.6\linewidth}
		\centering
		\hspace{-0.5cm}
		\includegraphics[width=1\linewidth]{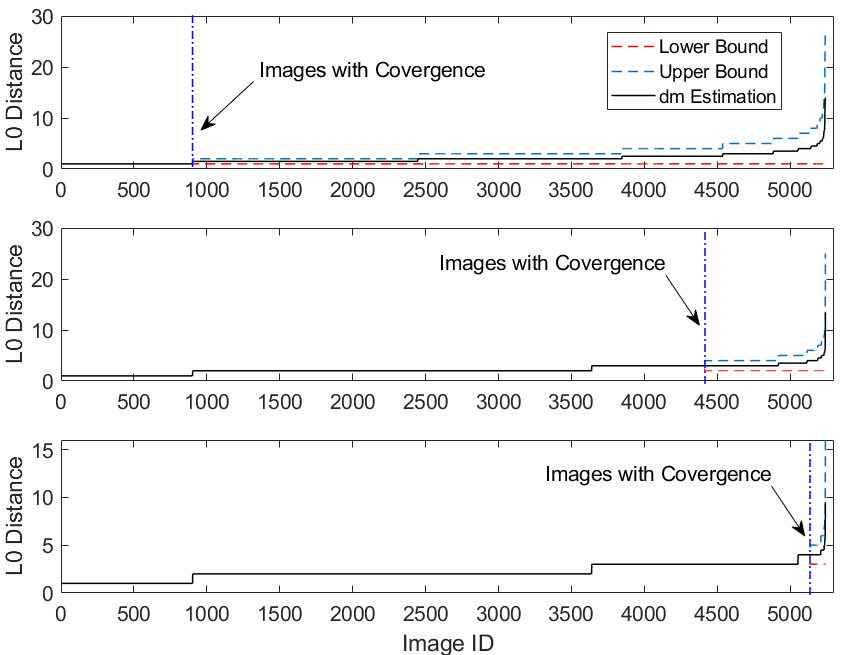}\\
		\text{(a)}
	\end{minipage}
	\begin{minipage}{0.4\linewidth}
		\centering
		\includegraphics[width=1.1\linewidth]{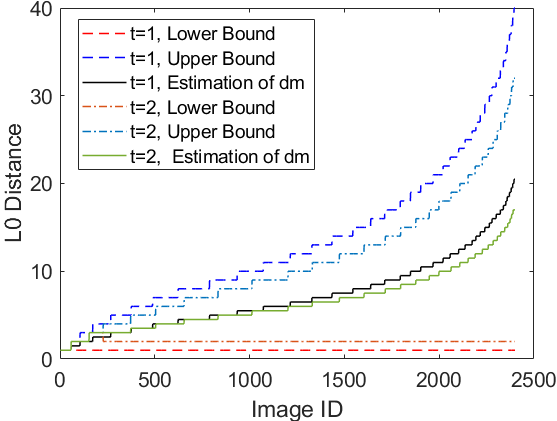}
		\text{(b)}
	\end{minipage}
		\vspace{-1mm}
	\caption{(a) sDNN: Upper bounds, lower bounds, and estimations of $d_m$ for all sampled images for $t \in \{ 1,2,3 \}$ ordered from the top to bottom. (b) DNN-0: Upper bounds, lower bounds, and estimations of $d_m$ for all sampled input images for $t \in \{ 1,2 \}$}
	\label{fig-2}
		\vspace{-2mm}
\end{figure}

We study the convergence and running time of our anytime global robustness evaluation algorithm on several DNNs in terms of the $L_0$-norm. To the best of our knowledge, no 
baseline method exists for this case study. The $L_0$-norm based algorithms,  which we compare against 
in Section~\ref{sec:competitiveL0}, cannot perform robustness evaluation based on both lower and upper bounds with provable guarantees.


We train two DNNs on the MNIST dataset, with DNN-0 being trained on the original images of size  $28\times 28$ and sDNN on images resized into $14\times 14$.
The model structures are given in Appendix \ref{sec:caseone}.
%
For DNN-0, we work with a set of 2,400 images randomly sampled from the dataset,
and for sDNN, we work with a set of 5,300 
images.



\subsubsection{sDNN: Speed of Convergence and Robustness Evaluation}

Fig.~\ref{fig-1}~(a) illustrates the speed of convergence of lower and upper bounds as well as the estimate for $d_m$ (i.e., the maximum safe radius) for an image with a large initial upper bound at $L_0$ distance~27. This image is chosen to demonstrate the \emph{worst case} for our approach. Working with a single image (i.e., local robustness) is the special case of our optimisation problem where $|\setofinputs|=1$.
We observe that, when transitioning from $t = 1$ to $t = 2$, the uncertainty radius $U_r(l_t,u_t)$ of $d_m$ 
is significantly reduced from 26 to 1, which demonstrates the effectiveness of our upper bound algorithm. Fig.~\ref{fig-1} (b) illustrates the speed of convergence of the global robustness evaluation on the testing dataset: Our method obtains tight upper and lower bounds efficiently and converges quickly. Notably, we have $U_c(l_t, u_t) = 1.97$ at $t=1$; the final global robustness is $2.1$, and thus, the relative error of the global robustness at $t=1$ is less than $7\%$. The estimate at $t=1$ can be obtained in polynomial time, and thus, our experimental results suggest that our approach provides a good approximation for this challenging NP-hard problem with reasonable error at very low computational cost.
%
%
Fig.~\ref{fig-1}~(c) gives the boxplots of the computational time required for individual iterations (\ie~subspace dimensions~$t$). We remark that at $t = 1$ it takes less than $0.1$\,s to process one image, which suggests that the algorithm has potential for real-time applications.

In Fig.~\ref{fig-2} (a), we plot the upper and lower bounds as well as the estimate for $d_m$ for all images in the testing dataset. The images are ordered using their upper bounds at $t = 1$. The dashed blue line indicates that all images left of this line have converged. The charts show a clear overall trend: our algorithm converges for most images with after a very small number of iterations.


\begin{figure}[t]
	\begin{minipage}{0.45\linewidth}
		\centering
		\includegraphics[width=1\linewidth]{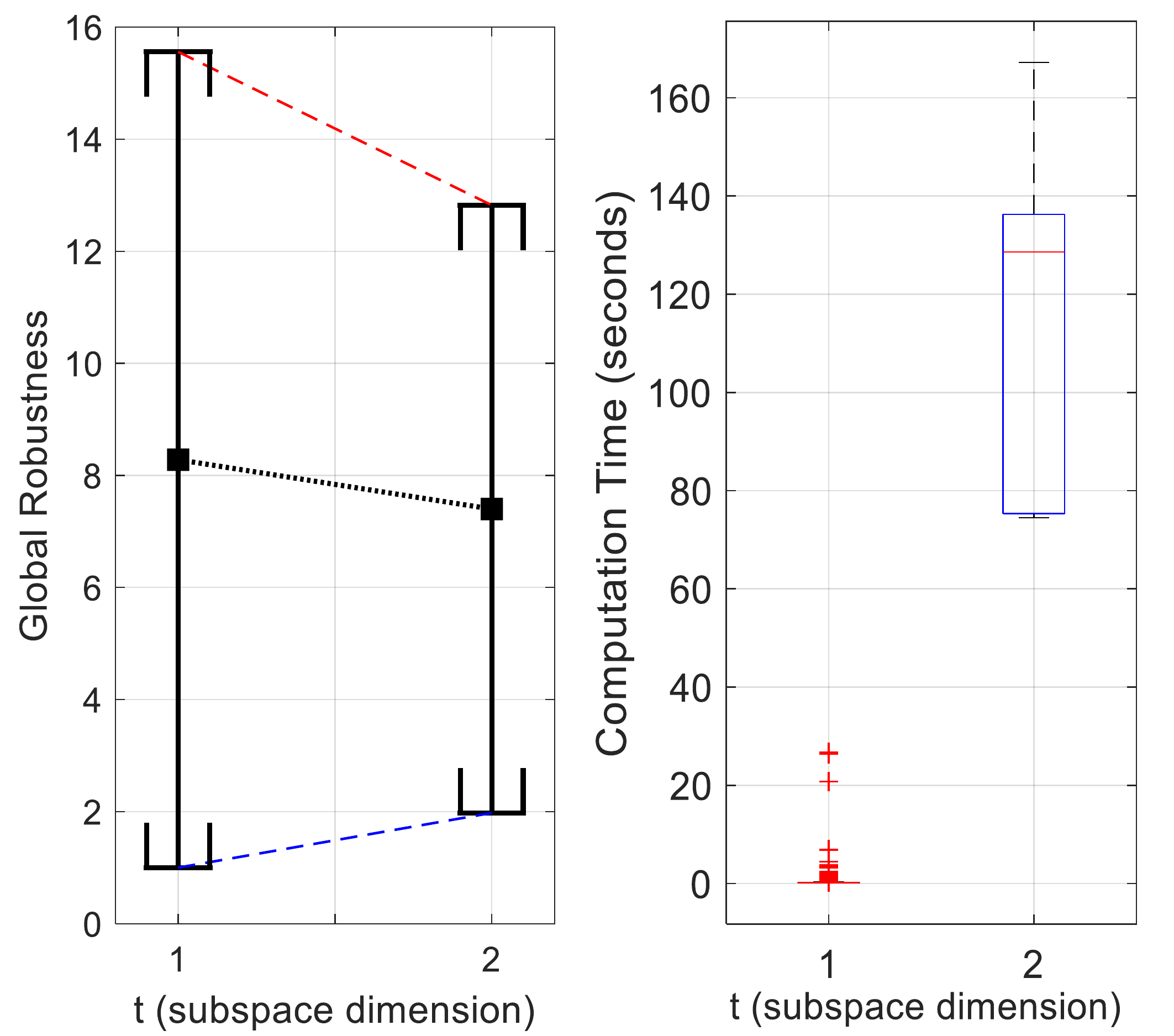}
		\caption{Robustness evaluation of DNN-0 for $t \in \{ 1, 2 \}$ and box-plots of computation time}
		\label{fig-3}
	\end{minipage}
	\hspace{0.5mm}
	\begin{minipage}{0.56\linewidth}
		\centering
		\includegraphics[width=1\linewidth]{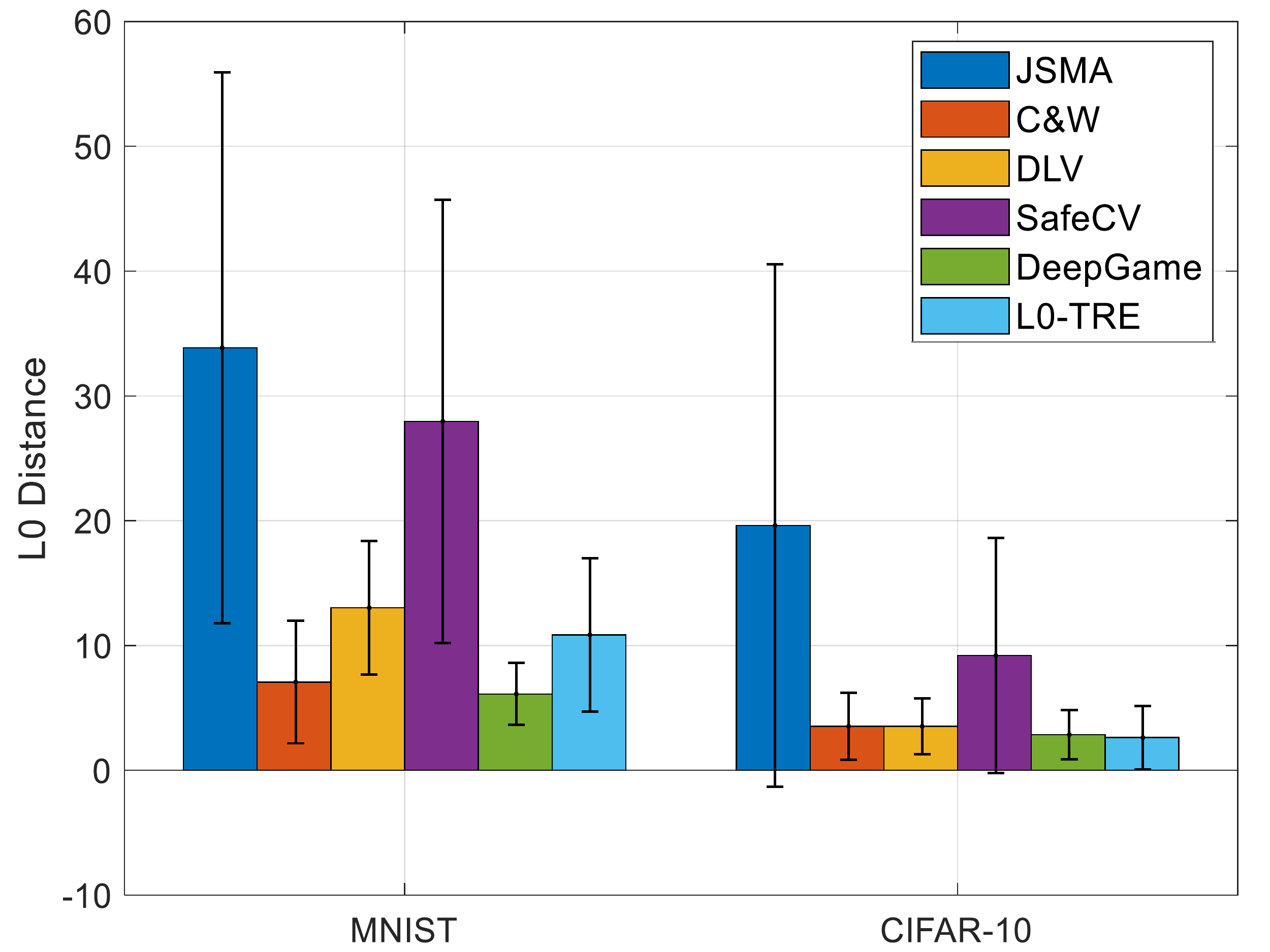}
		\caption{Means and standard deviations of the adversarial $L_0$ distance}
		\label{fig-6}
	\end{minipage}
		\vspace{-2mm}
\end{figure}
\begin{figure}[t]
		\begin{minipage}{0.51\linewidth}
		\centering
		\includegraphics[width=1.0\linewidth]{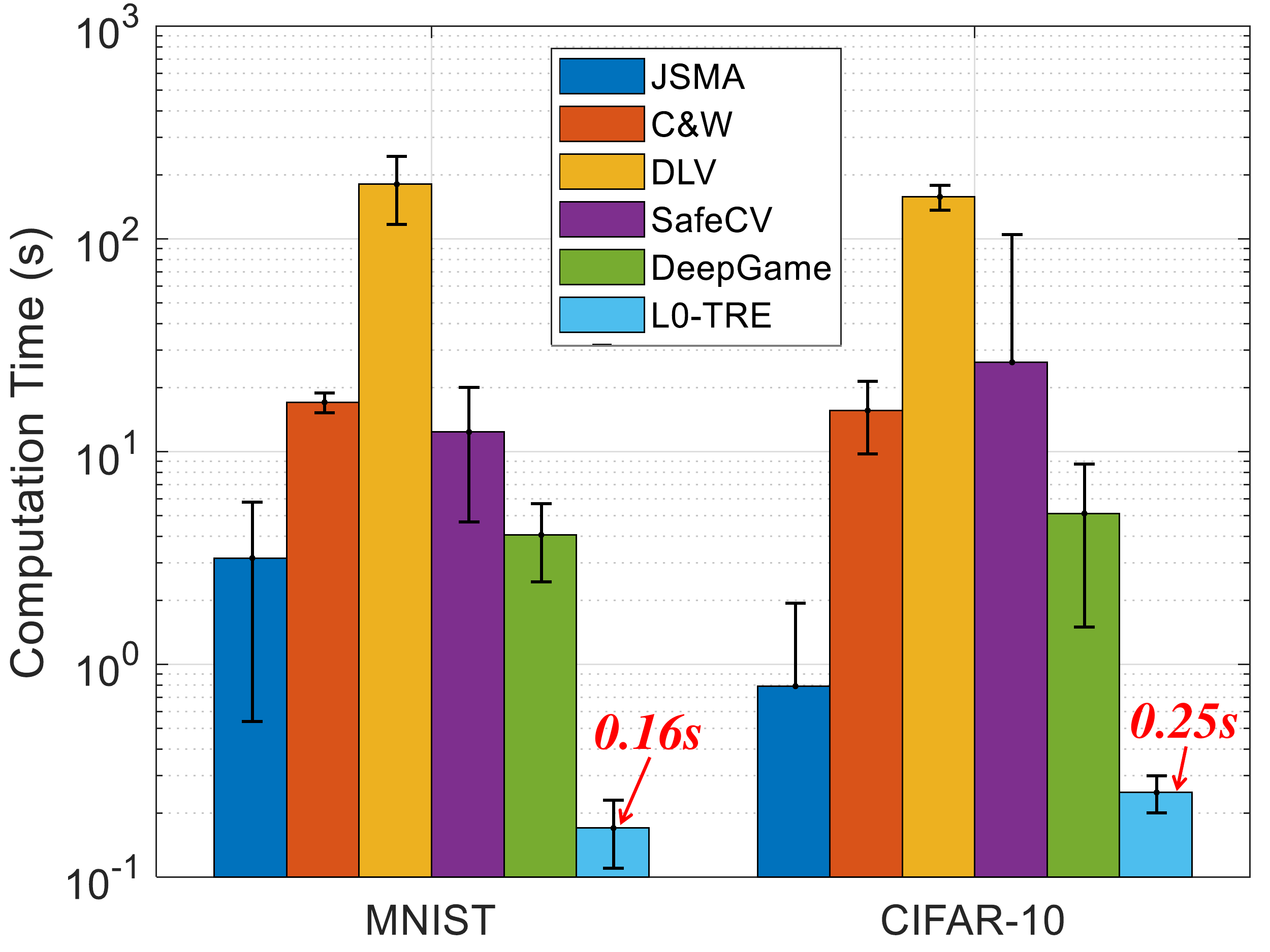}
		\caption{Means and standard deviations of \\computational time of all methods}
		\label{fig-7}
	\end{minipage}
		\hspace{0.5mm}
		\begin{minipage}{0.5\linewidth}
		\centering
		\includegraphics[width=1\linewidth]{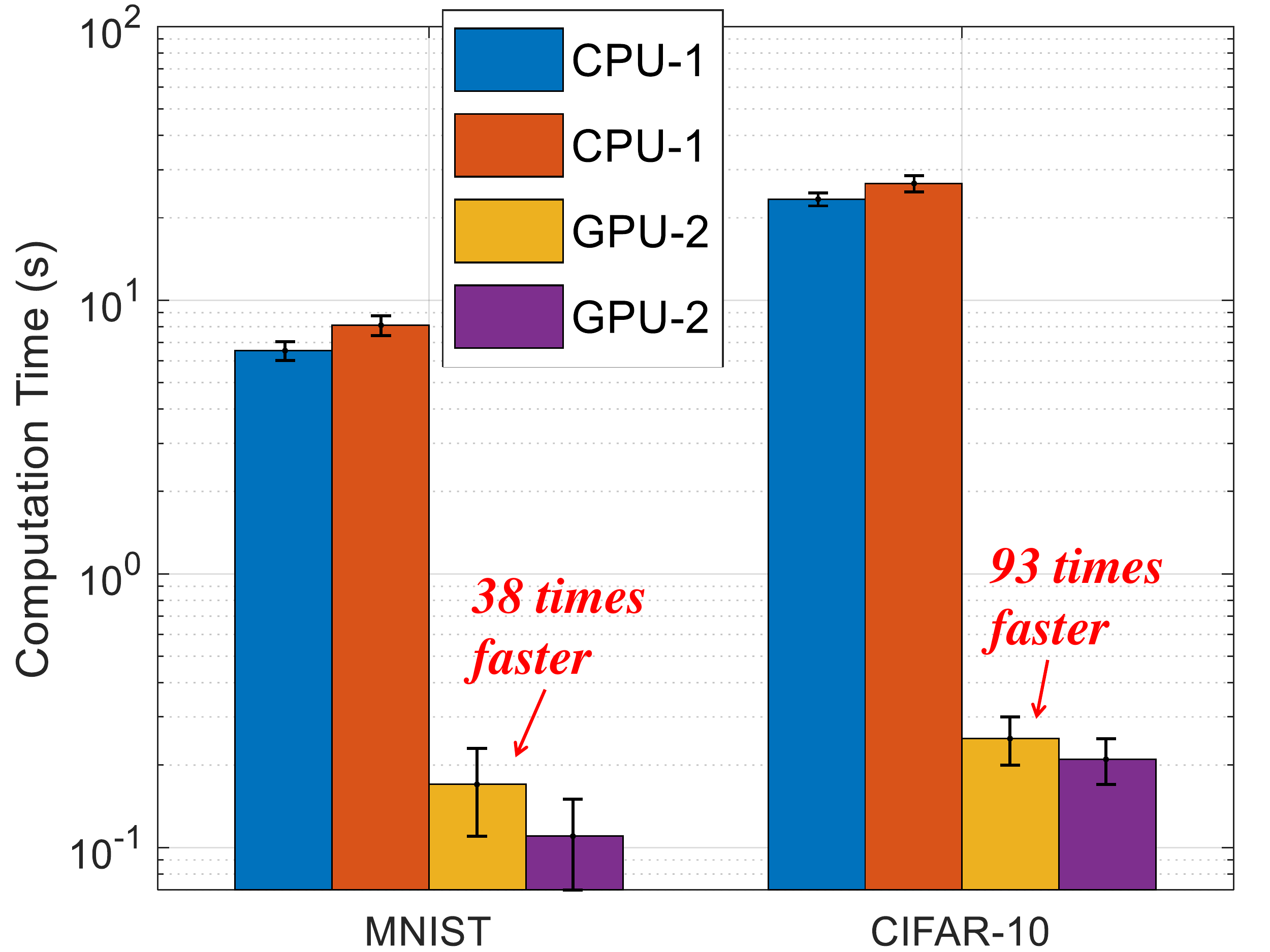}
		\caption{Means and standard derivations of computational time in CPU and GPU for 1,000 MNIST and CIFAR-10 images}
		\label{fig-5}
	\end{minipage}
		\vspace{-3mm}
\end{figure}

\subsubsection{DNN-0: Global Robustness Evaluation} 

Fig.~\ref{fig-2} (b) illustrates the overall convergence trends for all 2,400 images for our large DNN. We observe that even for a DNN with tens of thousands of hidden neurons, {$\TRE$} achieves tight estimates for $d_m$ for most images. Fig.~\ref{fig-3} gives the results of anytime global robustness evaluation at $t = 1$ and $t = 2$ for DNN-0. 
The results show the feasibility and efficiency of our approach for anytime global robustness evaluation of safety-critical systems. Fig.~\ref{fig-A2} in Appendix~\ref{sec:caseone} features a selection of the ground-truth adversarial images\footnote{Ground-true adversarial images mean images that are at the boundary of a safe norm ball, which is first proposed in~\cite{carlini2017ground}.} returned by our upper bound algorithm. 


\subsection{Case Study Two: $L_0$ Attacks}\label{sec:competitiveL0}



While the generation of attacks is not the primary goal of our method, we observe that our upper bound generation method is highly competitive with state-of-the-art methods for the computation of adversarial images. We train MNIST and CIFAR-10 DNNs and compare with JSMA~\cite{papernot2016limitations}, C\&W~\cite{carlini2017towards}, DLV~\cite{HKWW2017}, SafeCV~\cite{WHK2017} and DeepGame~\cite{wu2018game}, on 1,000 testing images. Details of the experimental settings are given in Appendix~\ref{sec:casethree}. 

\subsubsection{Adversarial $L_0$ Distance}
Fig.~\ref{fig-6} depicts the average and standard deviations of $L_0$ distances of the adversarial images produced by the five methods. A smaller $L_0$ distance indicates an adversarial example closer to the original image. For MNIST, the performance of our method is better than JSMA, DLV, and SafeCV, and comparable to C\&W and DeepGame. For CIFAR-10, the bar-chart reveals that our {$\TRE$} achieves the smallest $L_0$ distance (modifying 2.62 pixels on average) among all competitors. For this experiment, we stop at $t = 1$ without performing further iterations. 

\subsubsection{Computational Cost}
Fig.~\ref{fig-7} (note log-scale) gives runtimes. Our tensor-based parallelisation method delivers extremely efficient attacks. For example, for MNIST, our method is $18\times$, $100\times$, $1050\times$, and $357\times$ faster than JSMA, C\&W, DLV, and SafeCV, respectively. Figure~\ref{fig-5} shows that the tensor-based parallelisation\footnote{{\em CPU-1}: Tensorflow (Python) on i5-4690S CPU; {\em GPU-1}: Tensorflow (Python) with parallelisation on NVIDIA GTX TITAN GPU.~~{\em CPU-2}: Deep Learning Toolbox (Matlab2018b) on i7-7700HQ CPU; {\em GPU-2}: Deep Learning Toolbox (Matlab2018b) with parallelisation on NVIDIA GTX-1050Ti GPU.} significantly improves the computational efficiency in terms of 38 times faster on MNIST DNN and 93 times faster on CIFAR-10 DNN.
Appendix~\ref{sec:casethree} compares some of the adversarial examples found by the five methods. The examples illustrate that the modification of one to three pixels suffices to trigger misclassification even in well-trained neural networks.


\subsection{Case Study Three: Test Case Generation for DNNs}
\label{sec:testing}


\begin{table}[t]
    \caption{Neuron coverage achieved by $\TRE$, DeepConcolic and DeepXplore}
    \label{tab:nc}
    \centering
    \def\arraystretch{1.2}
    \scalebox{0.9}{
    \begin{tabular}{@{\quad}l@{\quad}|@{\quad}c@{\quad}|@{\quad}cccc@{\quad}}
    \toprule
         & \textbf{L0-TRE}  & DeepConcolic~\cite{swrhkk2018} & \multicolumn{3}{@{\quad}|@{\quad}c}{DeepXplore~\cite{deepxplore}~~(\%)}   \\ 
         & (\%) & (\%) & \multicolumn{1}{@{\quad}|@{\quad}c}{light} & occlusion & blackout \\ \hline
        MNIST & \textbf{98.95} & 97.60 & \multicolumn{1}{@{\quad}|@{\quad}c}{80.77} & 82.68 & 81.61 \\
        CIFAR-10  & \textbf{98.63} & 84.98 & \multicolumn{1}{@{\quad}|@{\quad}c}{77.56} &81.48 & 83.25 \\
    \bottomrule
    \end{tabular}
    }
\end{table}
\begin{figure}[t!]
    \centering
    \includegraphics[width=1\linewidth]{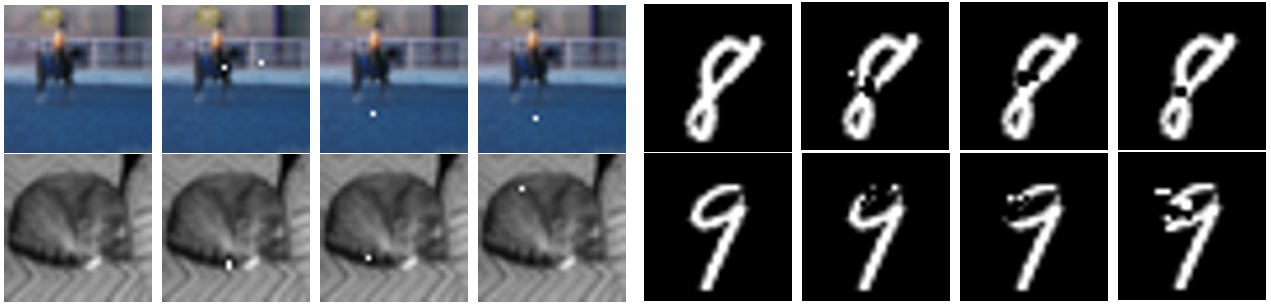}
    \caption{Some adversarial examples found by our tool $\TRE$ while generating test cases for high neuron coverage on MNIST and CIFAR-10 DNNs}
    \label{fig:Adversaries}
    	\vspace{-2mm}
\end{figure} 

A variety of methods to automate testing of DNNs has been proposed recently~\cite{deepxplore, sun2018testing, swrhkk2018}. The most widely used metric for the exhaustiveness of test suites for DNNs is \emph{neuron coverage}~\cite{deepxplore}. Neuron coverage quantifies the percentage of hidden neurons in the network that are activated at least once. We use $ne$ to range over hidden neurons, and $V(ne, x)$ to denote the activation value of $ne$ for test input~$x$. Then $V(ne, x)>0$ implies that $ne$ is covered by the test input $x$.

The application of our algorithm to coverage-driven test case generation is straight forward; it only requires a minor modification to the optimisation problem in Definition~\ref{def:localrobustness}. Given any neuron $ne$ that is not activated by the test suite $\setofinputs_0$, we find the input with the smallest distance to any input in $\setofinputs_0$ that activates $ne$. We replace the constraint $\classlabel(f,x_i) \neq \classlabel(f,x_{0,i})$ in Equation (\ref{eq:robustness}) with
\begin{equation}
V(ne, x_i)\leq 0 \wedge V(ne,x_{0,i})>0.
\end{equation}
The optimisation problem now searches for new inputs that activate the neuron $ne$, and the objective is to minimise the distance from the current set of test inputs $\setofinputs_0$.

We compare our tool $\TRE$ with other state-of-the-art test case generation methods, including DeepConcolic\footnote{Our optimisation algorithm has been also adopted in the testing tool \emph{DeepConcolic}, see \url{https://github.com/TrustAI/DeepConcolic}}~\cite{swrhkk2018} and DeepXplore~\cite{deepxplore}. All results are averaged over 10 runs or more. Table~\ref{tab:nc} gives the neuron coverage obtained by the three tools. We~observe that $\TRE$ yields much higher neuron coverage than both DeepConcolic and DeepXplore in any of its three modes of operation (`light', `occlusion', and `blackout'). Fig.~\ref{fig:Adversaries} depicts adversarial examples generated due to $L_0$ norm change by our tool $\TRE$ on test case generation. We also observe that a significant portion of the adversarial examples can be found using a relatively small $L_0$ distance. 
More experimental results can be found in Appendix \ref{sec:casefive}. 
Overall, our tool $\TRE$ offers an efficient approach to coverage-driven testing on DNNs.

Moreover, our tool can be used to guide the design of robust DNN architectures, as shown in Case Study Four (see Appendix~\ref{app:case4}). In Case Study Five, we show that $\TRE$ can also generate saliency map for model interpretability and is capable of evaluating local robustness for large-scale, state-of-the-art ImageNet DNN models including VGG-16/19, ResNet-50/101 and AlexNet (see Appendix~\ref{app:case5}).

\section{Related Work}
\label{sec:RW}


\subsection{Generation of Adversarial Examples}


Existing algorithms compute an upper bound of the maximum safety radius. However, they cannot guarantee to reach the maximum safety radius, while our method is able to produce both lower and upper bounds that provably converge to the maximum safety radius.
Most existing algorithms first compute a gradient (either a cost gradient or a forward gradient) and then perturb the input in different ways along the most promising direction on that gradient. 
\emph{FGSM} (Fast Gradient Sign Method)~\cite{goodfellow2014explaining} is for the $L_\infty$ norm. It computes the gradient $\nabla_{X}J(\theta, x,f(x))$. 
\emph{JSMA} (Jacobian Saliency Map based Attack)~\cite{papernot2016limitations} is for the $L_0$ norm. It calculates the Jacobian matrix of the output of a DNN (in the logit layer) with respect to the input. Then it iteratively modifies one or two pixels until a misclassification occurs. 
The \emph{C\&W Attack} (Carlini and Wagner)~\cite{carlini2017towards} works for the $L_0$, $L_2$ and $L_\infty$ norms. It formulates the search for an adversarial example as an image distance minimisation problem. The basic idea is to introduce a new optimisation variable to avoid box constraints (image pixels need to lie within $[0,1]$). 
\emph{DeepFool}~\cite{moosavi2016deepfool} works for the $L_2$ norm. It iteratively linearises the network around the input $x$ and moves across the boundary by a minimal step until reaching a misclassification. 
\emph{VAT} (Visual Adversarial Training)~\cite{miyato2015distributional} defines a KL-divergence at an input based on the model's robustness to the local perturbation of the input, and then perturbs the input according to this KL-divergence. 
We focus on the $L_0$~norm. We have shown experimentally that for this norm, our approach dominates all existing approaches. We obtain tighter upper bounds at lower computational cost.

\subsection{Safety Verification and Reachability Analysis}

The approaches 
aim to not only find an upper bound but also provide guarantees on the obtained 
bound. There are two ways of achieving safety verification for DNNs. The first is to reduce the problem to a constraint solving problem. Notable works include, e.g., \cite{PT2010,katz2017reluplex}. However, 
they can only work with small networks that have hundreds of hidden neurons. The second is to discretise the vector spaces of the input or hidden layers, and then apply exhaustive search algorithms or Monte-Carlo tree search algorithm on the discretised spaces. The guarantees are achieved by establishing local assumptions such as minimality of manipulations in \cite{HKWW2017} and minimum confidence gap for Lipschitz networks in \cite{WHK2017, wu2018game}. Moreover, \cite{LM2017} considers 
determining if an output value of a DNN is reachable from a given input subspace, and reduces the problem to a MILP problem; and
\cite{dutta2017output} considers the range of output values from a given input subspace. 
Both approaches can only work with small networks. We also mention \cite{PRGS2017}, which computes a lower bound of 
local robustness for the $L_2$ norm by propagating relations between layers backward from the output. It is incomparable with ours because of the different distance metrics. The bound is loose and 
cannot be improved (i.e., no convergence). 
Recently, some researchers use abstract interpretation to verify the correctness of DNNs~\cite{dd2018ai2,mirman2018differentiable}. Its basic idea is to use abstract domains (represented as e.g., boxes, zonotopes, polyhedra) to over-approximate the computation of a set of inputs. In recent work~\cite{gopinath2017deepsafe} the input vector space is partitioned using clustering and then the method of~\cite{katz2017reluplex} is used to check the individual partitions. DeepGO~\cite{RHK2018,RHK2018arXiv} shows that most known layers of DNNs are Lipschitz continuous and presents a verification approach based on global optimisation.

However, none of the verification tools above are workable on $L_0$-norm distance in terms of providing the \textit{anytime} and \textit{guaranteed} convergence to the true global robustness. Thus, the proposed tool, $\TRE$, is a supplementary to existing research on safety verification of DNNs.


\section{Conclusions}

In this paper, to evaluate global robustness of a DNN over a testing dataset, we present an approach to iteratively generate its lower and upper bounds. We show that the bounds are gradually, and strictly, improved and eventually converge to the optimal value. The method is  anytime, tensor-based, and offers provable guarantees. We conduct experiments on a set of challenging problems to validate our approach.


\bibliographystyle{splncs04}
\bibliography{L0GlobalRobust}

\newpage
\appendix

\section{Appendix: Proofs of Theorems}
\label{sec:proof}

\subsection{Proof of NP-hardness}
\label{sec:np}

\begin{mytheorem}
    Let $f: \mathbb{R}^n\to \mathbb{R}^m$ be a neural network and its input is normalized into $[0,1]^n$. When $\distance{D}{\cdot}$ is the $L_0$ norm, $d_m(f,x_0,\distance{D}{\cdot})$ is NP-hard, and there at least exists a deterministic algorithm that can compute  $d_m(f,x_0,\distance{D}{\cdot})$ in time complexity $O((1+1/\epsilon)^n)$ for the worst case scenario when the error tolerance for each dimension is $\epsilon > 0$.
\end{mytheorem}

\begin{myproof}
    Here we consider the worst case scenario and use a straight-forward grid search to verify the time complexity needed. In the worst case, the maximum radius of a safe $L_0$-norm ball for DNN $f$ is $d_m(f,x_0,\distance{D}{\cdot}) = n$. A grid search with grid size $\Delta = 1/\epsilon$ starts from $d_{L_0} = 1$ to verify whether $d_{L_0}$ is the radius of maximum safe $L_0$-norm ball and would require the following running time in terms of evaluation numbers of the DNN.
    \begin{equation}
    \sum_{d_{L_0}=1}^{n}\binom n{d_{L_0}}\Delta^{d_{L_0}} = (1+1/\epsilon)^n
    \end{equation}
\end{myproof}

From the above proof, we get the following remark.
\begin{myremark}
Computing $d_m(f,x_0,\distance{D}{\cdot})$ is {\em more challenging} problem for $D = 0$, since it requires a {\em higher} computing complexity than $D = 1$ and $D=2$. Namely, grid search only requires $(1/\epsilon)^n$ evaluations on DNN to estimate $d_m(f,x_0,\distance{1}{\cdot})$ or $d_m(f,x_0,\distance{2}{\cdot})$ given the same error tolerance $\epsilon$. 
\end{myremark}

\commentout{
\subsection{Guarantee of the Global Minimum by Grid Search}
\label{sec:proof-0}

We prove the following theorem that shows that grid search with $\epsilon$ can guarantee to find the global optimum given a certain error bound assuming that the neural network satisfies the Lipschitz condition, as adopted in~\cite{WHK2017,PRGS2017}.

\begin{mytheorem}[Guarantee of the global minimum of grid search]\label{proof-0}
	Assume a neural network $f(x): {[0,1]}^n \to \mathbb{R}^m$ is Lipschitz continuous w.r.t. a norm metric $||*||_D$ and its Lipschitz constant is $K$. By 
	recursively sampling $\Delta = 1/\epsilon$ in each dimension, denoted as $\mathcal{X} = \{x_1,\ldots,x_{\Delta^n}\}$, the following relation holds: 
	
	$$
	||f_{opt}(x^*) - min_{x\in\mathcal{X}}f(x)||_D \leq K||\dfrac{\epsilon}{2}\mathbf{I}_n||_D
	$$
	where $f_{opt}(x*)$ represents the global minimum value, $min_{x\in\mathcal{X}}f(x)$ denotes the minimum value returned by grid search, and $\mathbf{I}_n \in \mathbb{R}^{n\times n}$ is an all-ones matrix.
\end{mytheorem}

\begin{myproof}
	Based on the Lipschitz continuity assumption of $f$, we have
	$$
	||f(x_1) - f(x_2)||_D \leq K||x_1 - x_2||_D
	$$
	Based on the grid search with $\epsilon$, we then have $\forall \tilde{x}\in [0,1]^n, \exists x \in \mathcal{X}$ such that $||x^*-x||_D \leq ||\dfrac{\epsilon}{2}\mathbf{I}_n||_D$, denoted as $\mathcal{X}(\tilde{x})$. Thus the theorem holds given the fact that we can always find  $\mathcal{X}(x^*)$ from the sampled set $\mathcal{X}$ for the global minimum $x^*$.
\end{myproof}
}

\subsection{Proof of Theorem: Guarantee of Lower Bounds}	
\label{sec:proof-1}

\begin{myproof}
	Our proof proceeds by contradiction. Let $l=l(f,x_0)$. Assume that there is another adversarial example $x_0'$ such that $t' = \distance{0}{x_0' - x_0}\leq l$ where $t'$ represents the number of perturbed pixels. By the definition of adversarial examples, there exists a subspace $X_k \in \mathbb{R}^{t'}$ such that $cl(f,\mathcal{S}(x_0,t')[:,k]) \neq cl(f,x_0)$. By $t'\leq  l$, we can find a subspace $Y_q \in \mathbb{R}^{l}$ such that $X_k \subset Y_q$. Thus we have $S(Y_q,l) \geq S(X_k,t')$. Moreover, by $S(Y_q,l)\leq S(Y_1,l)$, we have $cl(f,\mathcal{S}(x_0,l)[:,1]) \neq cl(f,x_0)$ since $cl(f,\mathcal{S}(x_0,l)[:,p]) \neq cl(f,x_0)$. However, this conflicts with $cl(f,\mathcal{S}(x_0,l)[:,1]) = cl(f,x_0)$, which can be obtained by the algorithm for computing lower bounds in Section 4.2.
\end{myproof}

\subsection{Proof of Theorem: Guarantee of Upper Bounds}	
\label{sec:proof-2}

\begin{myproof}[Monotonic Decrease Property of Upper Bounds]
	We use mathematical induction to prove that upper bounds monotonically decrease.
	
	\textbf{Base Case $m = 1$}: Based on the algorithm in {\em Upper Bounds} of Section 4.2, we assume that, after $m=1$ subspace perturbations, we find the adversarial example $x'$ such that $cl(f,x') \neq cl(f,x_0)$. 
	
	We know that, at $t = i$, based on the algorithm, we get  $S (x_0,i)$ and $\mathcal{S}(x_0,i)$, the ordered subspace sensitivities and their corresponding subspaces. Assume that the ordered subspace list is $\{\phi_1,\phi_2,\ldots,\phi_m \}$. Then, from the assumption, we have $cl(f,x(\phi_1))\neq cl(f,x_0)$ where $x(\phi_1)$ denotes the input of the neural network corresponding to subspace~$\phi_1$. 
	
	Then, at $t = i+1$, according to the algorithm, we calculate $S (x_0,i+1)$ and $\mathcal{S}(x_0,i+1)$. Similarly, we assume the ordered subspace list is $\{\theta_1,\theta_2,\ldots,\theta_n\}$. Thus we can find a subspace $\theta_q$ in $\{\theta_1,\theta_2,\ldots,\theta_n\}$ such that $\phi_1 \subset \theta_q$. As a result, we know that $S(\phi_1)\leq S(\theta_q)$, thus $cl(f,x(\theta_q))\neq cl(f,x_0)$. After  exhaustive tightening, we can at least find its subset $\phi_1$, since $cl(f,\phi_1)\neq cl(f,x_0)$ still holds after removing the pixels $x = \theta_q - \phi_1$. So we know $||x(\theta_q) - x_0||_0\leq ||x(\phi_1) - x_0||_0$. However, $\theta_q$ will not necessarily be found at $t = i+1$ in our upper bound algorithm, depending on its location in $\{\theta_1,\theta_2,\ldots,\theta_n\}$:
	
	1. If it is in the front position such as $\{\theta_q,\theta_2,\ldots,\theta_m \}$, then we know that $||x(\theta_q) - x_0||_0\leq ||x(\phi_1) - x_0||_0$ based on the above analysis, \ie~$u_i(f,x_0) \geq u_{i+1}(f,x_0)$ holds.
	
	2. If it is in the behind position such as $\{\theta_1,\theta_2,\theta_q,\ldots,\theta_m \}$, we know that $S(\theta_1)\geq S(\theta_q) \geq S(\phi_1)$, which means that subspace $\theta_1$ leads to a larger network's confidence decrease than subspace $\theta_q$, thus $||x(\theta_1) - x_0||_0\leq ||x(\theta_q) - x_0||_0$. We already know $||x(\theta_q) - x_0||_0\leq ||x(\phi_1) - x_0||_0$, thus $u_i(f,x_0) \geq u_{i+1}(f,x_0)$ also holds.
	
	\textbf{Inductive Case $m = k$}: Assume that at $t =i$, after going through $m = k$ subspace perturbations, \ie~$\Phi_k = \{\phi_1 \cup \phi_2 \cup \ldots \cup \phi_k\}$, we find an adversarial example $x'$ and $u_i(f,x_0) \geq u_{i+1}(f,x_0)$ holds. We need to show that after going through $m = k+1$  subspace perturbations, \ie~$\Phi_{k+1} = \{\phi_1 \cup \phi_2 \cup \ldots \cup \phi_k  \cup \phi_{k+1}\}$, the relation $u_i(f,x_0) \geq u_{i+1}(f,x_0)$ also holds.
	
	Similarly, at $t = i+1$, we can find a $k$ subspace set $\Theta_{k} = \{\theta_{q_1} \cup \theta_{q_2} \cup \ldots \cup \theta_{q_k}\}$ such that $\Phi_k \subset \Theta_k$, and we know that $u_{i+1}(f,x_0) = ||x(\Phi_k) - x_0||_0 \leq ||x(\Theta_k)-x_0||_0=u_{i}(f,x_0)$ holds. Then, for the new subspace $\phi_{k+1}$ at $t = i$, we can also find $\theta_{q_{k+1}}$ at $t = i+1$ such that $\phi_{k+1} \subset \theta_{q_{k+1}}$. We get $\Theta_{k+1} =  \Theta_{k} \cup \theta_{q_{k+1}}$. As a result, we still have $\Phi_{k+1} \subset \Theta_{k+1}$, obviously, $cl(x(\Theta_{k+1}),f) \neq cl(x(\Phi_{k+1}),f)$, which means we can definitely find an adversarial example after all perturbations in $\Theta_{k+1}$. After exhaustive tightening process, we can at least find its subset $\Phi_{k+1}$, since $cl(f,x(\Phi_{k+1}))\neq cl(f, x_0)$ still holds after removing those pixels $x = \Theta_k - \phi_k$. So we know $||x(\Theta_{k+1}) - x_0||_0\leq ||x(\Phi_{k+1}) - x_0||_0$, \ie~$u_i(f,x_0) \geq u_{i+1}(f,x_0)$ holds.
\end{myproof}

\subsection{Proof of Theorem: Uncertainty Radius Convergence to Zero}	
\label{sec:proof-3}

\begin{myproof}[Uncertainty Radius Convergence to Zero: $\lim\limits_{t\to n}U_r(l_i,u_i) = 0$]
	Based on 
	the definition of $L_0$-norm distance (\ie~$0\leq l_i\leq d_m\leq u_i\leq n$), we know that  $t \rightarrow n \implies l_n \rightarrow n \implies U_r(l_i, u_i) = 1/2(u_i-l_i) = n - n = 0$.
\end{myproof}

\newpage
\clearpage

\section{Appendix: The $L_0$ Norm}\label{sec:l0distance}

\commentout{
The distance metric $\distance{D}{\cdot}$ can be any functional mapping $\distance{D}{\cdot}: \mathbb{R}^n \times \mathbb{R}^n \rightarrow [0, \infty]$ that satisfies the metric conditions. In this paper, we focus on the $L_0$ metric. For two inputs $x_0$ and $x$, their $L_0$ distance, denoted as $\distance{0}{x-x_0}$, is the number of elements that are different. When working with testing datasets, we define 
\begin{equation}
\begin{array}{lcll}
\distance{0}{\setofinputs-\setofinputs_0} & = & \expectation{x_0\in \setofinputs_0}{\distance{0}{x-x_0}} & \text{(our definition)}\\
& = & \frac{1}{|\setofinputs_0|}\sum_{x_0\in \setofinputs_0}\distance{0}{x-x_0} & \text{(all inputs in $\setofinputs_0$ are i.i.d.)}
\end{array}
\end{equation}
where $x$ is a homogeneous input of $x_0$ in $\setofinputs$.
While other norms such as $L_1$, $L_2$ and $L_\infty$ have been widely applied, see e.g., \cite{papernot2016limitations,kurakin2016adversarial}, in generating adversarial examples, the studies based  on the $L_0$ norm 
are few and far between. 
}

We justify on the basis of several aspects that the $L_0$ norm is worthy of  being considered.

\subsubsection{Technical Reason}

The reason why norms other than the $L_0$ norm are widely used is \emph{mainly technical}: the existing adversarial example generation algorithms \cite{papernot2016limitations,carlini2017towards} proceed by first computing the gradient $\nabla_x J(\theta,x,f(x))$, where $J(\theta,x,f(x))$ is a loss or cost function and $\theta$ are the learnable parameters of the network $f$, 
and then adapting (in different ways for different algorithms) the input $x$ into $x'$ along the gradient descent direction. To enable this computation, the change to the input $x$ needs to be \emph{continuous and differentiable}. It is not hard to see that, while the $L_1$, $L_2$ and $L_\infty$ norms are continuous and differentiable, the  $L_0$ norm is not. Nevertheless, the $L_0$ norm is an effective and efficient method to quantify a range of adversarial perturbations and should be studied.


\subsubsection{Tolerance of Human Perception to $L_0$ Norm}

Recently, \cite{papernot2016limitations} demonstrates through in-situ human experiments that the $L_0$ norm is good at approximating human visual perception. Specifically, 349 human participants were recruited for a study of visual perception on $L_0$ image distortions, with the result concluding that nearly all participants can correctly recognise $L_0$ perturbed images when the rate of distortion pixels is less than 5.61\% (i.e., 44 pixels for MNIST and 57 pixels for CIFAR-10) and 90\% of them can still recognise them when the distortion rate is less than 14.29\% (i.e., 112 pixels for MNIST and 146 pixels for CIFAR-10). 
This experiment essentially demonstrates that human perception is tolerant of perturbations with only a few pixels changed, and shows
the necessity of robustness evaluation based on the $L_0$  norm. 


\subsubsection{Usefulness of Approaches without Gradient}

From the security point of view, an attacker to a network may not be able to access its architecture and parameters, to say nothing of the gradient $\nabla_x J(\theta,x,f(x))$. Therefore, to evaluate the robustness of a network, we need to consider \emph{black-box} attackers, which can only query the network for classification. For a black-box attacker, an attack (or a perturbation) based on the $L_0$ norm is to change several pixels, which is arguably easier to initiate than attacks based on other norms, which often 
require modifications to nearly all the pixels.


\subsubsection{Effectiveness of Pixel-based Perturbations}

Perturbations by minimising the number of pixels to be changed have been shown to be effective. For example, \cite{WHK2017,SVS2017} show that manipulating a single pixel is sufficient for the classification to be changed for several networks trained on the CIFAR10 dataset 
and the Nexar traffic light challenge.
Our approach can beat the state-of-the-art pixel based perturbation algorithms  by finding tighter upper bounds to the maximum safety radius. As far as we know, this is the first work on finding lower bounds to the maximum safety radius.


\newpage
\clearpage

\section{Appendix: Discussion of Application Scenarios}
\label{sec:overview}
We now summarise 
possible application scenarios for the method proposed in this paper.

\subsubsection{Safety Verification}

Safety verification \cite{HKWW2017} is to determine, for a given network $f$, an input $x_0$, a distance metric $\distance{D}{\cdot}$, and a number $d$, whether the norm ball $X(f,x_0,\distance{D}{\cdot},d)$ is safe. Our approach will compute a sequence of lower bounds $\mathcal{L}(x_0)$ and upper bounds $\mathcal{U}(x_0)$ for the maximum safe radius $d_m(x_0)$. For every round $i>0$, we can claim one of the following cases: 
\begin{itemize}
	\item the norm ball $X(f,x_0,\distance{D}{\cdot},d)$ is safe when $d \leq \mathcal{L}(x_0)_i$ 
	\item the norm ball $X(f,x_0,\distance{D}{\cdot},d)$ is unsafe when $d \geq \mathcal{U}(x_0)_i$ 
	\item the safety of $X(f,x_0,\distance{D}{\cdot},d)$ is unknown when $\mathcal{L}(x_0)_i < d < \mathcal{U}(x_0)_i$. 
\end{itemize}
As a byproduct, our method can return at least one adversarial image for the second case. 

\subsubsection{Competitive $L_0$ Attack}

We have shown that the upper bounds in $\mathcal{U}(x_0)$ are monotonically decreasing. As a result, the generation of upper bounds 
can serve as a competitive $L_0$ attack method. 

\subsubsection{Global Robustness Evaluation}

Our method can have an asymptotic convergence to the true global robustness with provable guarantees. As a result, for two neural networks $f_1$ and $f_2$ that are trained for the same task (\eg~MNIST, CIFAR-10 or ImageNet) but with different parameters or architectures (\eg~different layer types, layer numbers or hidden neuron numbers), if $R(f_1,\distance{0}{\cdot}) > R(f_2,\distance{0}{\cdot})$ then we can claim that network $f_1$ is more robust than $f_2$ in terms of its resistance to $L_0$-norm adversarial attacks.

\subsubsection{Test Case Generation}

Recently software coverage testing techniques have been applied to DNNs and
several test criteria have been proposed, see e.g.,  
\cite{deepxplore,tian2017deeptest}.
Each criterion defines a set of requirements that have to be tested for a DNN.
Given a test suite, the coverage level of the set of requirements indicates the adequacy level for testing the DNN.
The technique in this paper can be conveniently  used for coverage-based testing of DNNs.



\subsubsection{Real-time Robustness Evaluation}

By replacing the exhaustive search in the algorithm 
with random sampling and formulating the subspace as a high-dimensional tensor (to enable  parallel computation with GPUs), our method becomes
\emph{real-time} 
(\eg~for a MNIST network, it takes around $0.1s$ to generate an adversarial example). A real-time evaluation can be useful for major safety-critical  applications, including self-driving cars, robotic navigation, \etc.
Moreover, our method can display in real-time a \emph{saliency map}, visualizing how classification decisions of the network are influenced by pixel-level sensitivities.

\newpage
\clearpage

\section{Appendix: Case Study Four: Guiding the Design of Robust DNN Architectures}
\label{app:case4}

In this case study, we show how to use our tool L0-TRE for guiding the design of robust DNN architectures.

\begin{figure}[ht]
		\centering
		\includegraphics[width=0.9\linewidth]{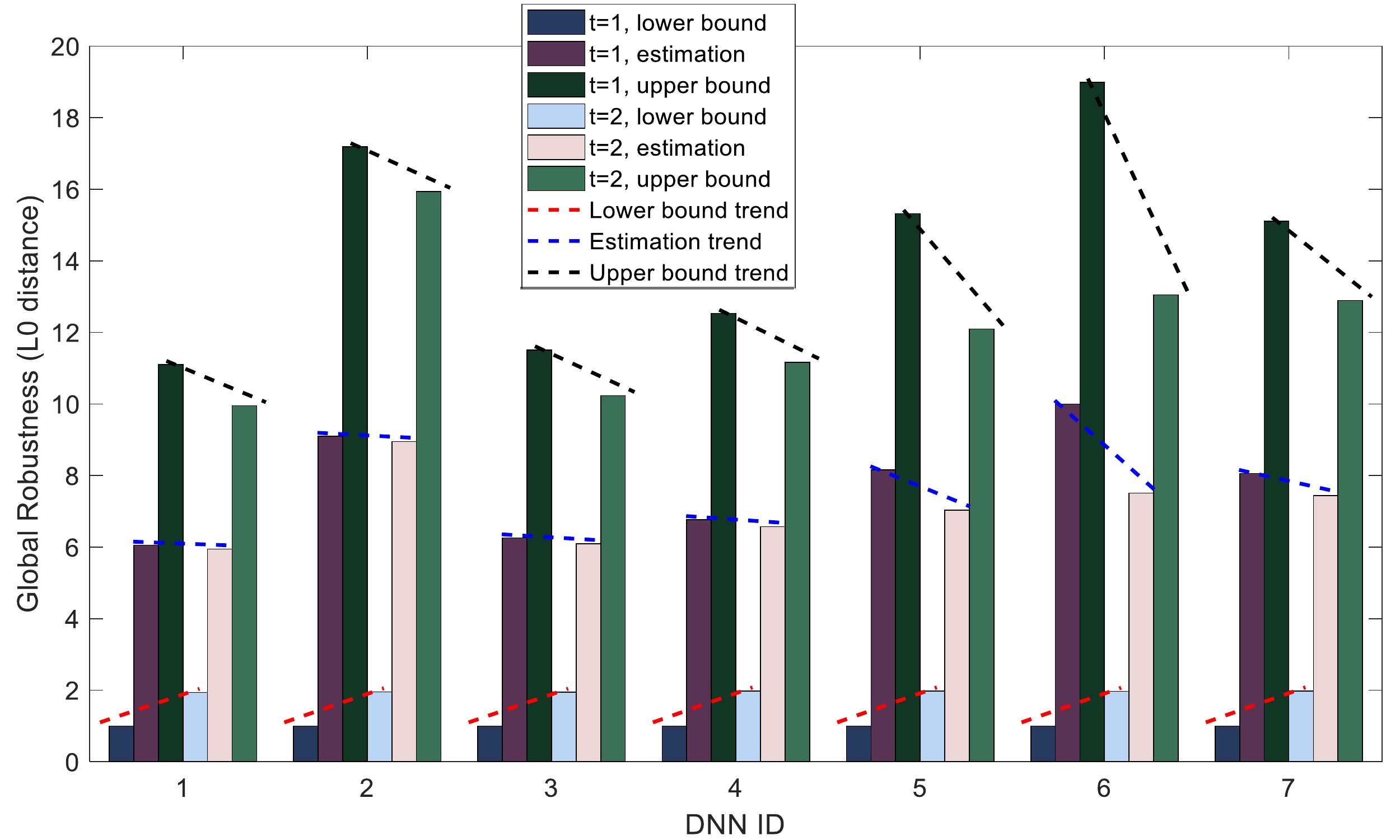}
		\caption{Lower bounds, upper bounds, and global robustness estimations for $t \in \{1, 2\}$ for seven DNN models}
		\label{fig-4app}
	\end{figure}

We trained seven DNNs, named DNN-$i$ for $i\in \{1,\ldots,7\}$, on the MNIST dataset using the same hardware and software platform and identical training parameters. The DNNs differ in their architecture, i.e., the number of layers and the types of the layers. The architecture matters: while all DNNs achieve 100\% accuracy during training, we observe accuracy down to 97.75\% on the testing dataset. Details of the models are in Appendix~\ref{sec:casetwo}. We aim to identify architectural choices that affect robustness.

\begin{figure}[t]
		\centering
		\includegraphics[width=0.6\linewidth]{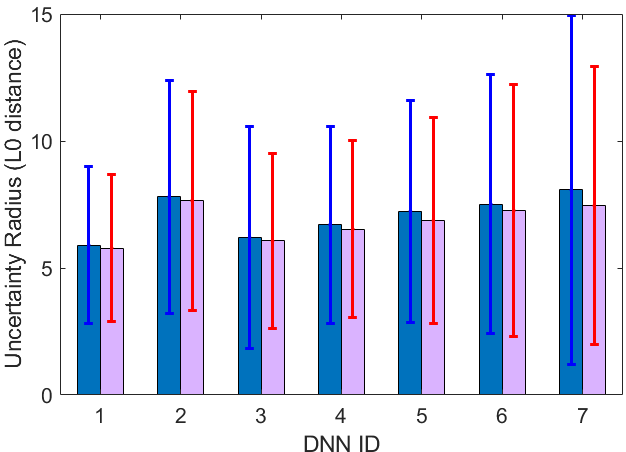}
	\caption{The means and standard derivations of $d_m$ uncertainty radiuses for 1,000 tested images at $t = 1,2$.}
	\label{fig-a1}
\end{figure}

Fig.~\ref{fig-4app} gives the estimates for global robustness, and their upper and lower bounds at $t = 1$ and $t=2$ for all seven DNNs. Fig.~\ref{fig-a1} illustrates the means and standard derivations of the $d_m$ estimates and the uncertainty radius for all 1,000 sampled testing images. And we also find that, the local robustness (\ie~robustness evaluated on a single image) of a network is coincident with its global robustness, so is the uncertainty radius. 

%
%
We observe the following:
\one~number of layers: a very deep DNN (\ie~too many layers relative to the size of the training dataset) is less robust, such as DNN-7;
%
\two~convolutional layers: DNNs with an excessive number of convolutional layers are less robust, e.g., compared with DNN-5, DNN-6 has an additional convolutional layer, but is significantly less robust; 
%
\three~batch-normalisation layers: adding a batch-normalisation layer may improve robustness, e.g., DNN-3 is more robust than DNN-2.

We remark that testing accuracy is not a good proxy for robustness: a DNN with higher testing accuracy is not necessarily more robust, e.g., DNN-1 and DNN-3 are more robust than DNN-6 and DNN-7, which have higher testing accuracies. DNNs may require a balance between robustness and their ability to generalise (proxied by testing accuracy). DNN-4 is a good example, among our limited set.

In summary, our tool L0-TRE can be used to choose what kind of layers should be added into a deep learning model for a safety-critical system, or given two neural networks with similar testing accuracies, which one should be chose, or finding a balanced model between robustness and performance.


%
%
%
%
%
%
%
%
%
%
%
%
%

\newpage
\clearpage

\section{Appendix: Case Study Five: Saliency Map and Local Robustness Evaluation for Large-scale ImageNet DNN Models}
\label{app:case5}


\subsection{Local Robustness Evaluation for ImageNet Models}

\begin{figure}[ht]
		\centering
		\includegraphics[width=0.7\linewidth]{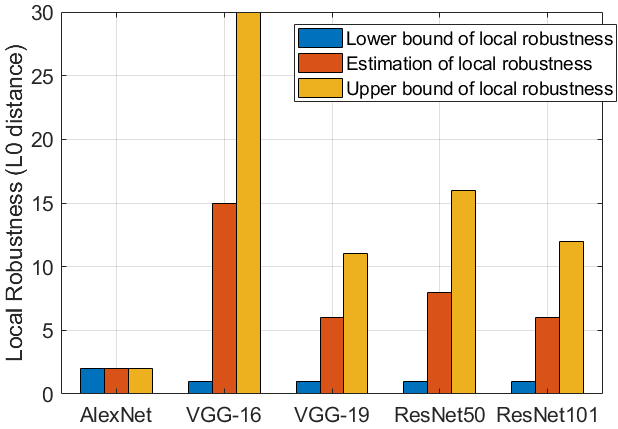}
		\caption{The upper bound, lower bound and estimation of local robustness for five ImageNet DNNs on a given input image}
		\label{fig:mean_std_ImageNet_DNNs}
\end{figure}

\begin{figure}[t]
		\centering
	\includegraphics[width=1\linewidth]{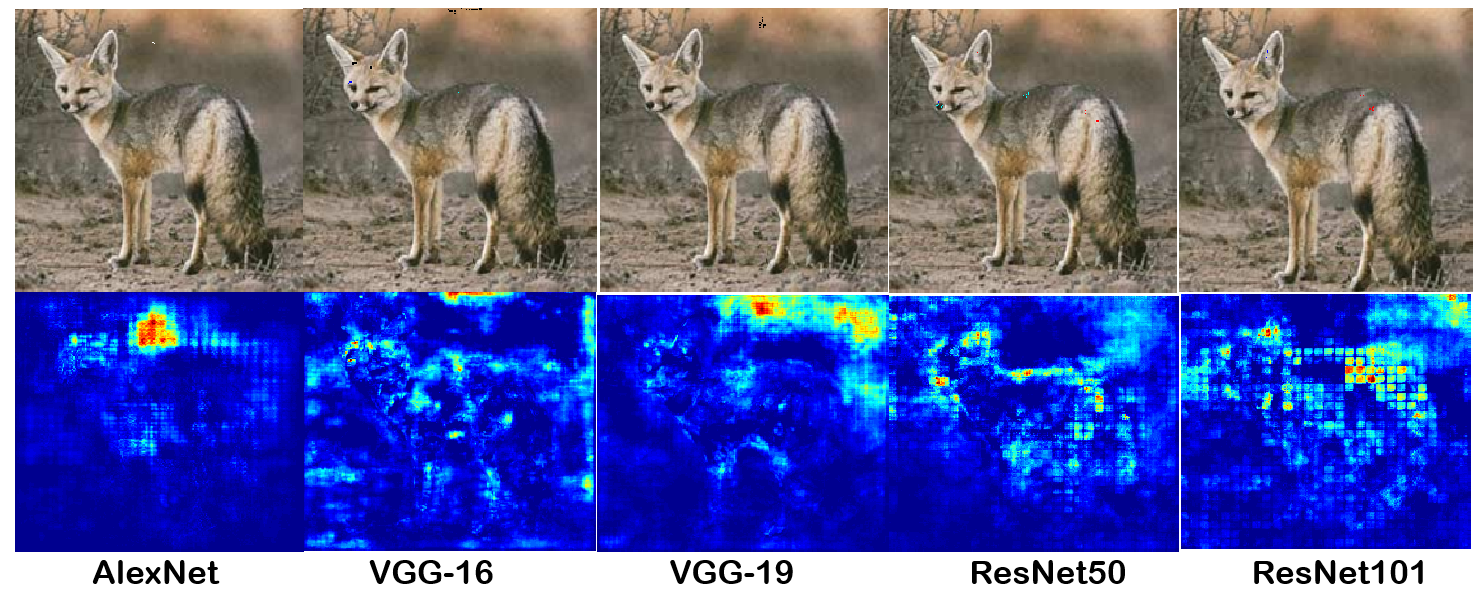}
	\caption{Adversarial examples on upper boundaries (top) and saliency maps (bottom)}
	\label{fig:saliency}
\end{figure}

We apply our method to five state-of-the-art ImageNet DNN models, including AlexNet (8 layers), VGG-16 (16 layers), VGG-19 (19 layers), ResNet50 (50 layers), and ResNet101 (101 layers). We set  $t = 1$ and 
generate the lower/upper bounds and estimates of local robustness for an input image. 
Fig.~\ref{fig:mean_std_ImageNet_DNNs} gives the local robustness estimates and their bounds for these networks. The adversarial images on the upper boundaries are featured in the top row of Fig.~\ref{fig:saliency}. For AlexNet, on this specific image, {L0-TRE} is able to find its ground-truth adversarial example (local robustness converges at $L_0$ = 2). We also observe that, for this image, the most robust model is VGG-16 (local robustness = 15) and the most vulnerable one is AlexNet (local robustness = 2). Fig.~\ref{fig:mean_std_ImageNet_DNNs} also reveals that, for similar network structures such as VGG-16 and VGG-19, ResNet50 and ResNet101, a model with deeper layers is less robust to adversarial perturbation. This observation is consistent with our conclusion in Case Study Three.

The method proposed in this paper, just as shown in this case study, provide a possible way to practically evaluate the robustness for large-scale DNN models and such robustness evaluation is with provable bounded guarantees. As a byproduct, our method can also generate saliency map for each input image as shown by the second column of Fig.~\ref{fig:saliency}.

\subsection{Saliency Map Generation}

Model interpretability (or explainability) addresses the problem that the decisions of DNNs are difficult to explain. Recent work, such as~\cite{LL2017}, calculates the contribution of each input dimension to the output decision. Our computation of subspace sensitivity $S(\setofinputs,t)$ can be re-used to quantify this contribution for each pixel of an image.

As shown in Fig.~\ref{fig:saliency} (more examples in Appendix~\ref{sec:casefour}), a brighter area indicates vulnerability to perturbation; it is very easy to see that VGG-16 is the most robust model. We obtain a large bright area for AlexNet, where a minor perturbation can lead to a misclassification. On the contrary, for this image, there are no vulnerable areas for VGG-16. The constraints of the optimisation problem given in Definition~\ref{def:localrobustness} can be adapted to generate full saliency maps for the hidden neurons, which have potential as an explanation for the decisions that DNNs make~\cite{olah2018the}. Moreover, a classic concept in cooperative game theory is to calculate the contribution of the players to their cooperation. Quantifying such contribution values play an important role in game-based safety verification on DNNs such as recent works in~\cite{WHK2017,wu2018game}. The intermediate result from L0-TRE tool in terms of subspace sensitivity $S(\setofinputs,t)$, as shown in this case study, is well suitable for this purpose as validated in paper~\cite{wu2018game}.

\newpage
\clearpage
\section{Appendix: Experimental Settings for Case Study One} \label{sec:caseone}

\subsection{Model Structures of sDNN}
\label{sec:caseone-1}

\begin{table}
	\caption{sDNN}
	\centering
	\begin{tabular}{ c | c }
		\toprule
		Layer Type & Size\\
		\hline
		Input layer & 14 $\times$ 14 $\times$ 1 \\ 
		Convolution layer & 2 $\times$ 2 $\times$ 8  \\ 
		Batch-Normalization layer & 8 channels  \\ 
		ReLU activation & \\
		Convolution layer & 2 $\times$ 2 $\times$ 16  \\ 
		Batch-Normalization layer & 16 channels  \\ 
		ReLU activation & \\
		Convolution layer & 2 $\times$ 2 $\times$ 32  \\ 
		Batch-Normalization layer &  32 channels \\ 
		ReLU activation & \\
		Fully Connected & 10  \\ 
		softmax + Class output &   \\ 
		\bottomrule    
	\end{tabular}
\end{table}

\subsection{Parameter Settings of sDNN}
\label{sec:caseone-2}
\subsubsection{Model Training Setup} 
\begin{itemize}
	\item Hardware: Notebook PC with I7-7700HQ, 16GB RAM, GTX 1050 GPU
	
	\item Software: Matlab 2018a, Neural Network Toolbox, Image Processing Toolbox, Parallel Computing Toolbox
	
	\item Parameter Optimization Settings: SGDM, Max Epochs = 20, Mini-Batch Size = 128
	
	\item Training Dataset: MNIST training dataset with 50,000 images
	
	\item Training Accuracy: 99.5\%
	
	\item Testing Dataset: MNIST testing dataset with 10,000 images
	
	\item Testing Accuracy: 98.73\%
\end{itemize}

\subsubsection{Algorithm Setup} 
\begin{itemize}
	\item $\epsilon = 0.25$
	
	\item Maximum $t = 3$
	
	\item Tested Images: 5,300 images sampled from MNIST testing dataset 
\end{itemize}

\subsection{Model Structures of DNN-0}
\label{sec:caseone-3}

\begin{table}
	\caption{DNN-0}
	\centering
	\begin{tabular}{ c | c }
		\toprule
		Layer Type & Size\\
		\hline
		Input layer & 28 $\times$ 28 $\times$ 1 \\ 
		Convolution layer & 3 $\times$ 3 $\times$ 32  \\ ReLU activation & \\
		Convolution layer & 3 $\times$ 3 $\times$ 64  \\ ReLU activation & \\
		Maxpooling layer & 2 $\times$ 2\\
		Dropout layer & 0.25\\
		Fully Connected layer & 128  \\
		ReLU activation & \\
		Dropout layer & 0.5\\
		Fully Connected layer & 10  \\ 
		Softmax + Class output &   \\ 
		\bottomrule    
	\end{tabular}
\end{table}

\subsection{Parameter Settings of DNN-0}
\label{sec:caseone-4}
\subsubsection{Model Training Setup} 
\begin{itemize}
	\item Hardware: Notebook PC with I7-7700HQ, 16GB RAM, GTX 1050 GPU
	
	\item Software: Matlab 2018a, Neural Network Toolbox, Image Processing Toolbox, Parallel Computing Toolbox
	
	\item Parameter Optimization Settings: SGDM, Max Epochs = 30, Mini-Batch Size = 128
	
	\item Training Dataset: MNIST training dataset with 50,000 images
	
	\item Training Accuracy: 100\%
	
	\item Testing Dataset: MNIST testing dataset with 10,000 images
	
	\item Testing Accuracy: 99.16\%
\end{itemize}

\subsubsection{Algorithm Setup} 
\begin{itemize}
	\item $\epsilon = 0.25$
	
	\item Maximum $t = 2$
	
	\item Tested Images: 2,400 images sampled from MNIST testing dataset 
\end{itemize}

\subsection{Ground-Truth Adversarial Images}
\label{sec:caseone-5}

\begin{figure}[t]
	\center
	\begin{minipage}{0.9\linewidth}
		\centering
		\includegraphics[width=1\linewidth]{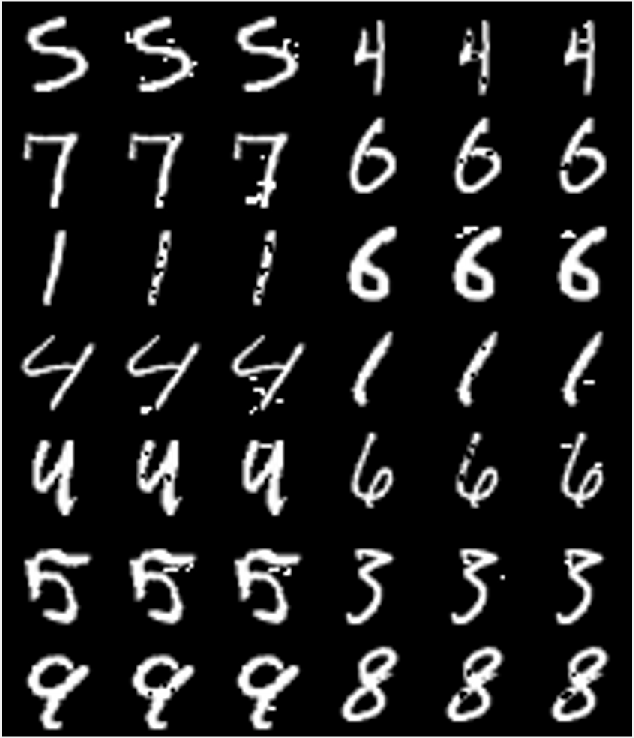}
	\end{minipage}
	\caption{Ground truth adversarial examples when converging to $d_m$. \textit{For each digital image, from the left to right, the first is original image, the second is the adversarial image returned at $t = 1$, and the third is the adversarial example at the boundary of a safe norm ball, namely the ground-truth adversarial examples~\cite{carlini2017ground}.}}
	\label{fig-A2}
\end{figure}

Fig.~\ref{fig-A2} displays some adversarial images returned by our upper bound algorithm.

\begin{figure}[t]
	\includegraphics[width=1\linewidth]{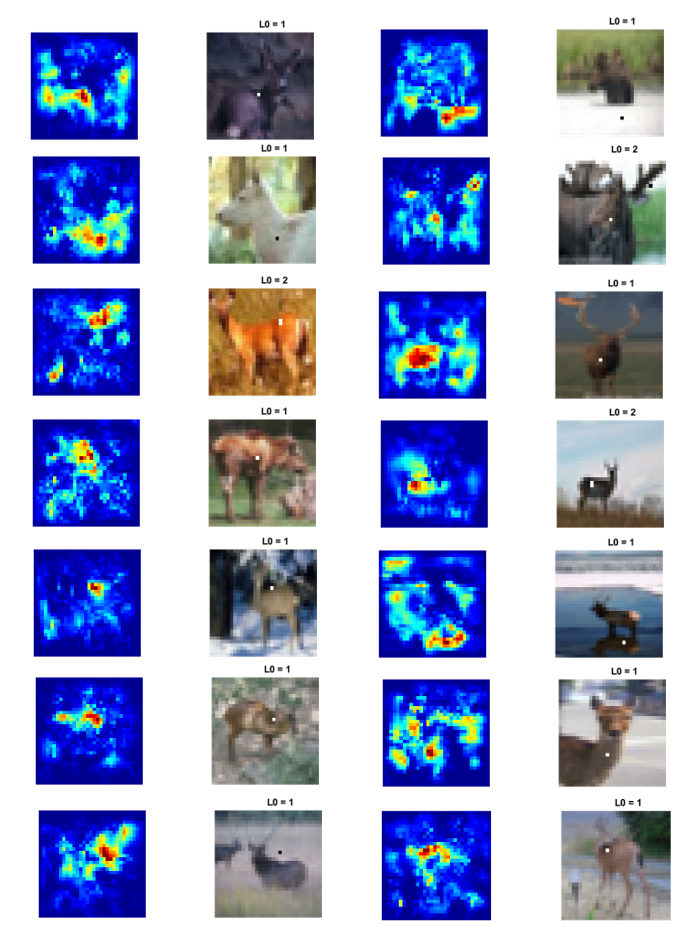}
	\caption{Ground-truth adversarial examples (right column) generated by {L0-TRE} at $t = 1$, and the saliency maps of the original images (left column).}
	\label{fig-A5}
\end{figure}

\newpage
 \clearpage 
\section{Appendix: Experimental Settings for Case Study Two} \label{sec:casethree}

\subsection{Model Structures for $L_0$ Attack}
\label{sec:casethree-1}
The architectures for the MNIST and CIFAR-10 models used in $ L_0 $ attack are illustrated in Table~\ref{tbl:case3-MNIST+CIFAR10}.

\begin{table}[h!]
	\caption{Model architectures for the MNIST and CIFAR-10 models.}
	\label{tbl:case3-MNIST+CIFAR10}
	\centering
	\begin{tabular}{ l | l | l }
		\toprule
		Layer Type & MNIST & CIFAR-10 \\
		\hline
		Convolution + ReLU & 3 $\times$ 3 $\times$ 32 & 3 $\times$ 3 $\times$ 64 \\ 
		Convolution + ReLU & 3 $\times$ 3 $\times$ 32 & 3 $\times$ 3 $\times$ 64 \\ 
		Max Pooling & 2 $\times$ 2 & 2 $\times$ 2 \\ 
		Convolution + ReLU & 3 $\times$ 3 $\times$ 64 & 3 $\times$ 3 $\times$ 128 \\ 
		Convolution + ReLU & 3 $\times$ 3 $\times$ 64 & 3 $\times$ 3 $\times$ 128 \\ 
		Max Pooling & 2 $\times$ 2 & 2 $\times$ 2 \\ 
		Flatten &  &  \\ 
		Fully Connected + ReLU & 200 & 256 \\ 
		Dropout & 0.5 & 0.5 \\ 
		Fully Connected + ReLU & 200 & 256 \\ 
		Fully Connected & 10 & 10 \\ 
		\bottomrule    
	\end{tabular}
\end{table}

\subsection{Model Training Setups}
\label{sec:casethree-2}
\begin{itemize}
	\item Parameter Optimization Option: Batch Size = 128, Epochs = 50, Loss Function = tf.nn.softmax\_cross\_entropy\_with\_logits, Optimizer = SGD(lr=0.01, decay=1e-6, momentum=0.9, nesterov=True)
	
	\item Training Accuracy: 
	\begin{itemize}
		\item MNIST (99.99\% on 60,000 images)
		\item CIFAR-10 (99.83\% on 50,000 images)
	\end{itemize}
	\item Testing Accuracy:
	\begin{itemize}
		\item MNIST (99.36\% on 10,000 images)
		\item CIFAR-10 (78.30\% on 10,000 images)
	\end{itemize}
\end{itemize}

\subsection{Adversarial Images}
\label{sec:casethree-5}

\begin{figure}[ht]
	\center
	\begin{minipage}{1.0\linewidth}
		\centering
		\includegraphics[width=0.95\linewidth]{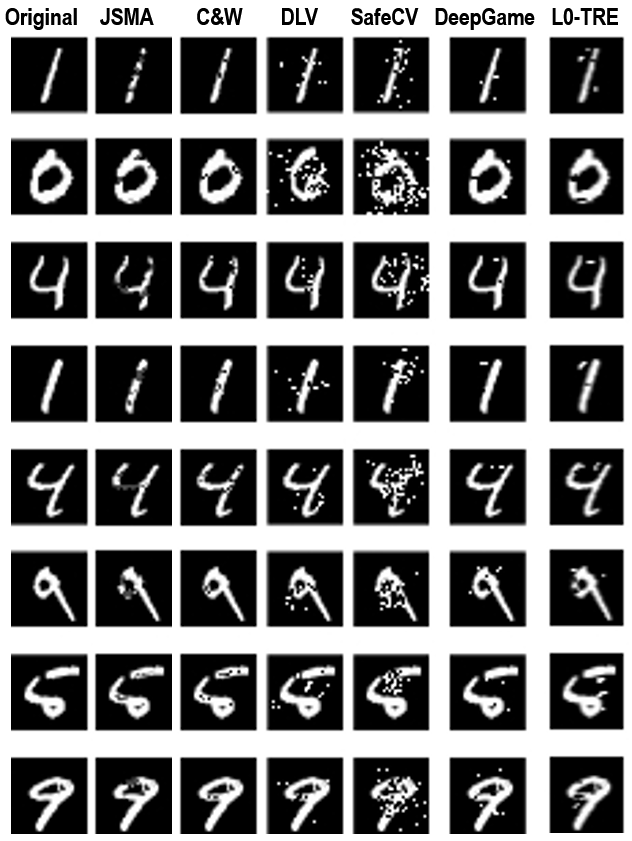}
	\end{minipage}
	\hspace{1mm}
	\caption{Adversarial images generated by the $L_0$ attack methods on the MNIST dataset. From left to right: original image, JSMA, C\&W, DLV, SafeCV, DeepGame, and our tool L0-TRE.}
	\label{fig-A3}
\end{figure}

\begin{figure}[ht]
	\center
	\begin{minipage}{1.0\linewidth}
		\centering
		\includegraphics[width=0.95\linewidth]{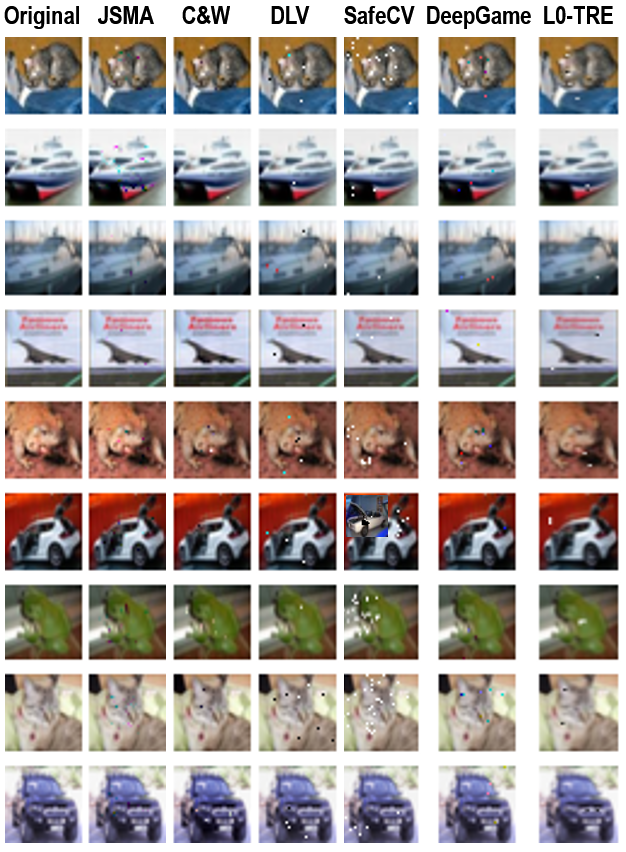}
	\end{minipage}
	\hspace{1mm}
	\caption{Adversarial images generated by the $L_0$ attack methods on the CIFAR-10 dataset. From left to right:original image, JSMA, C\&W, DLV, SafeCV, DeepGame, and our tool L0-TRE.}
	\label{fig-A4}
\end{figure}

Fig.~\ref{fig-A3} and~\ref{fig-A4} present a few adversarial examples generated on the MNIST and CIFAR-10 datasets by our approach {L0-TRE}, together with results for four other tools, \ie~C\&W, JSMA, DLV, and SafeCV.

\subsection{Experimental Setting for Competitive L0 Attack Comparison}\label{sec:casethree-3}

\subsubsection{Baseline Methods}
We choose four well-established baseline methods that can perform state-of-the-art $L_0$ adversarial attacks. Their code is available on GitHub.

\begin{itemize}
	\item JSMA\footnote{\url{https://github.com/tensorflow/cleverhans/blob/master/cleverhans_tutorials/mnist_tutorial_jsma.py}}: is a targeted attack based on the $L_0$-norm, 
	here used so that adversarial examples are misclassified into all classes except the correct one.
	
	\item C\&W\footnote{\url{https://github.com/carlini/nn_robust_attacks}}: is a state-of-the-art adversarial attack method, which models the attack problem as an unconstrained optimization problem that is solvable by the Adam optimizer in Tensorflow.
	
	\item DLV\footnote{\url{https://github.com/TrustAI/DLV}}: is an untargeted DNN verification method based on exhaustive search and Monte Carlo tree search (MCTS).
	
	\item SafeCV\footnote{\url{https://github.com/matthewwicker/SafeCV}}: is a feature-guided black-box safety verification and attack method based on the Scale Invariant Feature Transform (SIFT) for feature extraction, game theory, and MCTS.
	
	\item DeepGame\footnote{\url{https://github.com/TrustAI/DeepGame}}: is a two-player turn-based game framework for the verification of deep neural networks with provable guarantees on $L_1$, $L_2$ and $L_\infty$-norm distances, but with a slight modification it can be used to perform $L_0$-norm adversarial attack. 
	
\end{itemize}

\subsubsection{Dataset} We perform comparison on two datasets - MNIST and CIFAR-10. They are standard benchmark datasets for adversarial attack of DNNs, and are widely adopted by all these baseline methods.

\begin{itemize}
	
	\item MNIST dataset\footnote{\url{http://yann.lecun.com/exdb/mnist/}}: is an image dataset of handwritten digits, which contains a training set of 60,000 examples and a test set of 10,000 examples. The digits have been size-normalized and centered in a fixed-size image.
	
	\item CIFAR-10 dataset\footnote{\url{https://www.cs.toronto.edu/~kriz/cifar.html}}: is an image dataset of 10 mutually exclusive classes, i.e., `airplane', `automobile', `bird', `cat', `deer', `dog', `frog', `horse', `ship', `truck'. It consists of 60,000 32x32 colour images - 50,000 for training, and 10,000 for testing. 
	
\end{itemize}

\subsubsection{Platforms}

\begin{itemize}
	
	\item Hardware Platform:
	\begin{itemize}
		\item NVIDIA GeForce GTX TITAN Black
		\item Intel(R) Core(TM) i5-4690S CPU @ 3.20GHz $ \times $ 4
	\end{itemize}
	
	\item Software Platform: 
	\begin{itemize}
		\item Ubuntu 14.04.3 LTS
		\item Fedora 26 (64-bit)
		\item Anaconda, PyCharm
	\end{itemize}
	
\end{itemize}

\subsection{Algorithm Settings}
\label{sec:casethree-4}

MNIST and CIFAR-10 use the same settings, unless separately specified.
\begin{itemize}
	\item JSMA: 
	\begin{itemize}
		\item bounds = (0, 1)
		\item predicts = `logits'
	\end{itemize}
	\item C\&W: 
	\begin{itemize}
		\item targeted = False
		\item learning\_rate = 0.1
		\item max\_iteration = 100
	\end{itemize}
	\item DLV: 
	\begin{itemize}
		\item mcts\_mode = ``sift\_twoPlayer"
		\item startLayer, maxLayer = -1
		\item numOfFeatures = 150
		\item featureDims = 1
		\item MCTS\_level\_maximal\_time = 30
		\item MCTS\_all\_maximal\_time = 120
		\item MCTS\_multi\_samples = 5 (MNIST), 3 (CIFAR-10)
	\end{itemize}
	
	\item SafeCV:
	\begin{itemize}
		\item MANIP = max\_manip (MNIST), white\_manipulation (CIFAR-10)
		\item VISIT\_CONSTANT = 1
		\item backtracking\_constant = 1
		\item simulation\_cutoff = 100
	\end{itemize}
	
    \item DeepGame
    \begin{itemize}
    \item gameType = `cooperative'
        \item bound = `ub'
        \item algorithm = `A*'
        \item eta = (`L0', 30)
        \item tau = 1
    \end{itemize}
	
	\item Ours: 
	\begin{itemize}
		\item EPSILON = 0.5
		\item L0\_UPPER\_BOUND = 100
	\end{itemize}
\end{itemize}

\newpage
 \clearpage 


\section{Appendix: Experimental Settings for Case Study Three} \label{sec:casefive}

The experimental settings of this case study can be found in Appendix~\ref{sec:casethree}.

\begin{figure}[ht]
	\vspace{-5mm}
		\centering
		\includegraphics[width=0.63\linewidth]{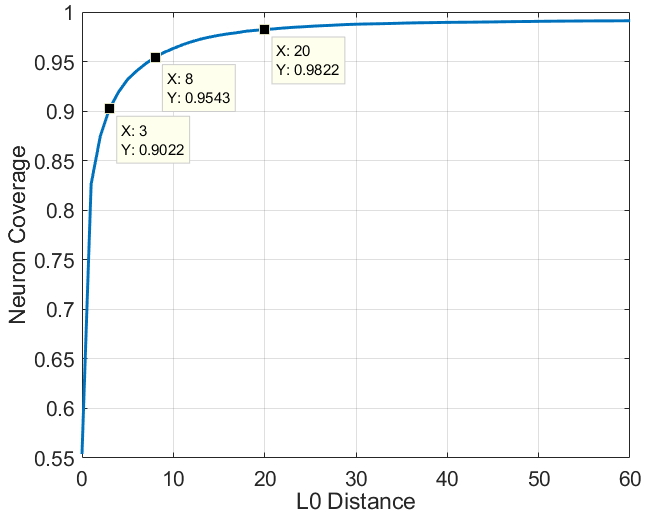}\\
		\text{(a) MNIST}\\
		\includegraphics[width=0.63\linewidth]{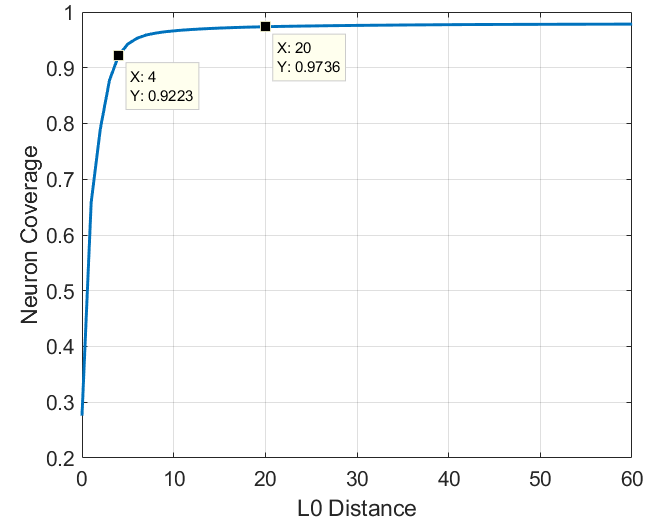}\\
	\text{(b) CIFAR-10}
	\caption{Neuron coverage by robustness evaluation on MNIST (a) and CIFAR-10 (b). \textit{The horizontal axis measures the $L_0$ distance of each generated test case with respect to the original input, and the vertical axis records the coverage percentage. We see that it achieves more than \%90 neuron coverage by only modifying 3 pixels.}}
	\label{fig:testing-nc}
\end{figure}

\begin{figure}[t]
	\centering
	\includegraphics[width=0.65\columnwidth]{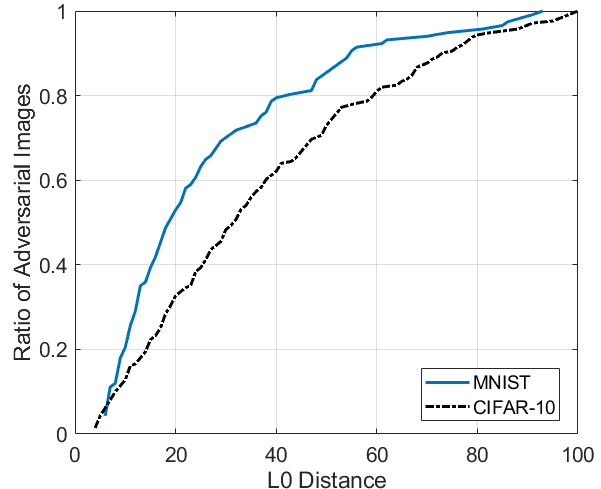}
	\caption{Neuron coverage: the percentage of adversarial examples within each distance. \textit{It depicts that a significant portion of adversarial examples can be found in the relatively small $L_0$ distance end of the curve.} More experimental results and applications of applying L0-TRE into DNN testing can be found in GitHub \url{https://github.com/TrustAI/DeepConcolic}}
	\label{fig:testing-advs}
\end{figure}

\newpage
\clearpage
\section{Appendix: Experimental Settings for Case Study Four} \label{sec:casetwo}

\subsection{Model Training and Algorithm Setup} 
\begin{itemize}
	\item Hardware: Notebook PC with I7-7700HQ, 16GB RAM, GTX 1050 GPU
	
	\item Software: Matlab 2018a, Neural Network Toolbox, Image Processing Toolbox, Parallel Computing Toolbox
	
	\item Parameter Optimization Settings: SGDM, Max Epochs = 30, Mini-Batch Size = 128
	
	\item Training Dataset: MNIST training dataset with 50,000 images
	
	\item Training Accuracy: All seven models reach 100\%
	
	\item Testing Dataset: MNIST testing dataset with 10,000 images
	
	\item Testing Accuracy:  DNN-1 = 97.75\%; DNN-2 = 97.95\%; DNN-3 = 98.38\%; DNN-4 = 99.06\%; DNN-5 = 99.16\%; DNN-6 = 99.13\%; DNN-7 = 99.41\%
	
	\item L0-TRE Algorithm Setup: $\epsilon = 0.3$, Maximum $t = 2$, Tested Images: 1,000 images sampled from MNIST testing dataset 
\end{itemize}

\subsection{Model Structures of DNN-1 to DNN-7}
\label{sec:casetwo-1}

The model structures of DNN-1 to DNN-7 are described in respective tables.

\begin{table}[ht]
	
	\begin{minipage}{0.5\linewidth}
		\centering
		\caption{DNN-1}
		\centering
		\begin{tabular}{ c | c }
			\toprule
			Layer Type & Size\\
			\hline
			Input layer & 28 $\times$ 28 $\times$ 1 \\ 
			Convolution layer & 3 $\times$ 3 $\times$ 32  \\ 
			ReLU activation & \\
			Fully Connected layer & 10  \\ 
			Softmax + Class output &   \\ 
			\bottomrule    
		\end{tabular}
	\end{minipage}
	\begin{minipage}{0.5\linewidth}
		\centering
		\caption{DNN-2}
		\centering
		\begin{tabular}{ c | c }
			\toprule
			Layer Type & Size\\
			\hline
			Input layer & 28 $\times$ 28 $\times$ 1 \\ 
			Convolution layer & 3 $\times$ 3 $\times$ 32  \\ 
			ReLU activation & \\
			Convolution layer & 3 $\times$ 3 $\times$ 64  \\ 
			ReLU activation & \\
			Fully Connected layer & 10  \\ 
			Softmax + Class output &   \\ 
			\bottomrule    
		\end{tabular}
	\end{minipage}
\end{table}

\begin{table}
	\begin{minipage}{0.5\linewidth}
		\centering
		\caption{DNN-3}
		\centering
		\begin{tabular}{ c | c }
			\toprule
			Layer Type & Size\\
			\hline
			Input layer & 28 $\times$ 28 $\times$ 1 \\ 
			Convolution layer & 3 $\times$ 3 $\times$ 32  \\ 
			ReLU activation & \\
			Convolution layer & 3 $\times$ 3 $\times$ 64  \\
			Batch-Normalization layer & \\
			ReLU activation & \\
			Fully Connected layer & 10  \\ 
			Softmax + Class output &   \\ 
			\bottomrule    
		\end{tabular}
	\end{minipage}
	\begin{minipage}{0.5\linewidth}
		\centering
		\caption{DNN-4}
		\centering
		\begin{tabular}{ c | c }
			\toprule
			Layer Type & Size\\
			\hline
			Input layer & 28 $\times$ 28 $\times$ 1 \\ 
			Convolution layer & 3 $\times$ 3 $\times$ 32  \\ 
			ReLU activation & \\
			Convolution layer & 3 $\times$ 3 $\times$ 64  \\
			Batch-Normalization layer & \\
			ReLU activation & \\
			Fully Connected layer & 128  \\ 
			ReLU activation & \\
			Fully Connected layer & 10  \\ 
			Softmax + Class output &   \\ 
			\bottomrule    
		\end{tabular}
	\end{minipage}
\end{table}

\begin{table}
	\begin{minipage}{0.5\linewidth}
		\centering
		\caption{DNN-5}
		\centering
		\begin{tabular}{ c | c }
			\toprule
			Layer Type & Size\\
			\hline
			Input layer & 28 $\times$ 28 $\times$ 1 \\ 
			Convolution layer & 3 $\times$ 3 $\times$ 32  \\ 
			ReLU activation & \\
			Convolution layer & 3 $\times$ 3 $\times$ 64  \\
			Batch-Normalization layer & \\
			ReLU activation & \\
			Dropout layer & 0.5 \\
			Fully Connected layer & 128  \\ 
			ReLU activation & \\
			Fully Connected layer & 10  \\ 
			Softmax + Class output &   \\ 
			\bottomrule    
		\end{tabular}
	\end{minipage}
	\begin{minipage}{0.5\linewidth}
		\centering
		\caption{DNN-6}
		\centering
		\begin{tabular}{ c | c }
			\toprule
			Layer Type & Size\\
			\hline
			Input layer & 28 $\times$ 28 $\times$ 1 \\ 
			Convolution layer & 3 $\times$ 3 $\times$ 32  \\ 
			ReLU activation & \\
			Convolution layer & 3 $\times$ 3 $\times$ 64  \\ 
			ReLU activation & \\
			Convolution layer & 3 $\times$ 3 $\times$ 128  \\
			Batch-Normalization layer & \\
			ReLU activation & \\
			Dropout layer & 0.5 \\
			Fully Connected layer & 128  \\ 
			ReLU activation & \\
			Fully Connected layer & 10  \\ 
			Softmax + Class output &   \\ 
			\bottomrule    
		\end{tabular}
	\end{minipage}
\end{table}

\begin{table}
	\caption{DNN-7}
	\centering
	\begin{tabular}{ c | c }
		\toprule
		Layer Type & Size\\
		\hline
		Input layer & 28 $\times$ 28 $\times$ 1 \\ 
		Convolution layer & 3 $\times$ 3 $\times$ 16  \\ 
		ReLU activation & \\
		Convolution layer & 3 $\times$ 3 $\times$ 32  \\
		Batch-Normalization layer & \\
		ReLU activation & \\
		Convolution layer & 3 $\times$ 3 $\times$ 64  \\
		Batch-Normalization layer & \\
		ReLU activation & \\
		Convolution layer & 3 $\times$ 3 $\times$ 128  \\
		Batch-Normalization layer & \\
		ReLU activation & \\
		Dropout layer & 0.5 \\
		Fully Connected layer & 256  \\ 
		ReLU activation & \\
		Dropout layer & 0.5 \\
		Fully Connected layer & 10  \\ 
		Softmax + Class output &   \\ 
		\bottomrule    
	\end{tabular}
\end{table}

\newpage
\clearpage
\section{Appendix: Experimental Settings for Case Study Five} \label{sec:casefour}

\subsection{State-of-the-art ImageNet DNN Models}
\label{sec:casefour-1}
\begin{itemize}
	\item AlexNet~\cite{krizhevsky2012imagenet}
	: a convolutional neural network, which was originally designed in the ImageNet Large Scale Visual Recognition Challenge in 2012. The network achieved a top-5 error of 15.3\%, more than 10.8 percentage points ahead of the runner up.
	
	\item VGG-16 and VGG-19~\cite{simonyan2014very}
	: were released in 2014 by the Visual Geometry Group at the University of Oxford. This family of architectures achieved second place in the 2014 ImageNet Classification competition, achieving 92.6\% top-five accuracy on the ImageNet 2012 competition dataset. 
	
	\item ResNet50 and ResNet101~\cite{he2016deep}
	: are designed based on residual nets with a depth of 50 and 101 layers. An ensemble of these networks achieved 3.57\% testing error on ImageNet and won the 1st place on the ILSVRC 2015 classification task. Their variants also won 1st places on the tasks of ImageNet detection, ImageNet localization, COCO detection, and COCO segmentation.
	
\end{itemize}

\subsection{Experimental Settings}
\label{sec:casefour-2}

\subsubsection{Platforms} 
\begin{itemize}
	\item Hardware: Notebook PC with I7-7700HQ, 16GB RAM, GTX 1050 GPU
	
	\item Software: Matlab 2018a, Neural Network Toolbox, Image Processing Toolbox, Parallel Computing Toolbox, and AlexNet, VGG-16, VGG-19, ResNet50 and ResNet101 Pretrained DNN Models
\end{itemize}

\subsubsection{Algorithm Setup} 
\begin{itemize}
	\item $\epsilon = 0.3$
	
	\item Maximum $t = 1$
	
	\item Tested Images: 20 ImageNet images, randomly chosen
\end{itemize}

\subsection{Adversarial Images and Saliency Maps}

\begin{figure}
	\includegraphics[width=0.75\linewidth]{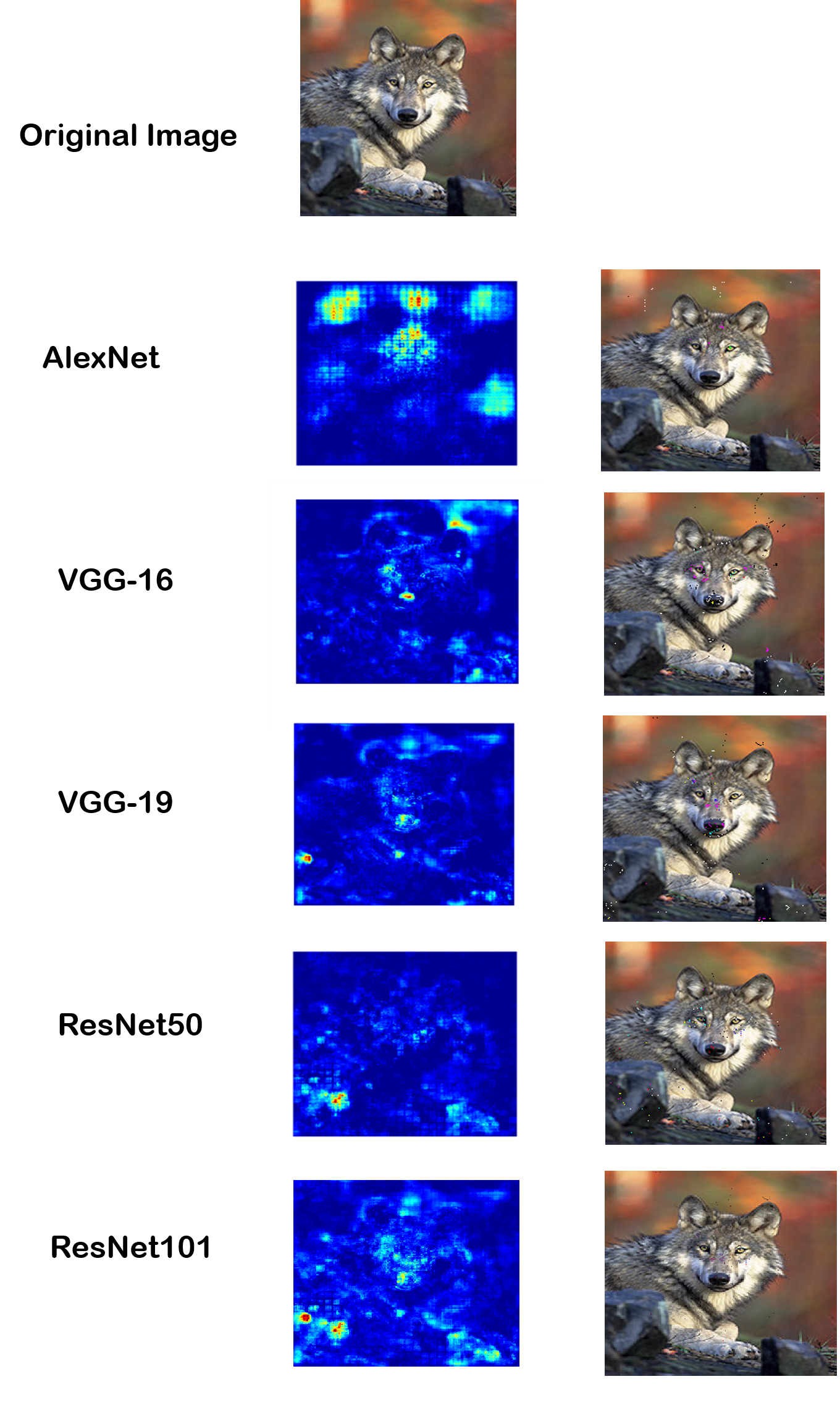}
	\caption{Adversarial examples on upper boundaries returned by our tool {L0-TRE} (right column), and saliency maps for each ImageNet DNN model (left column).}
	\label{fig:casefour-3}
\end{figure}

Fig.~\ref{fig:casefour-3} gives more examples of adversarial images and saliency maps. 

\begin{figure}
	\includegraphics[width=1\linewidth]{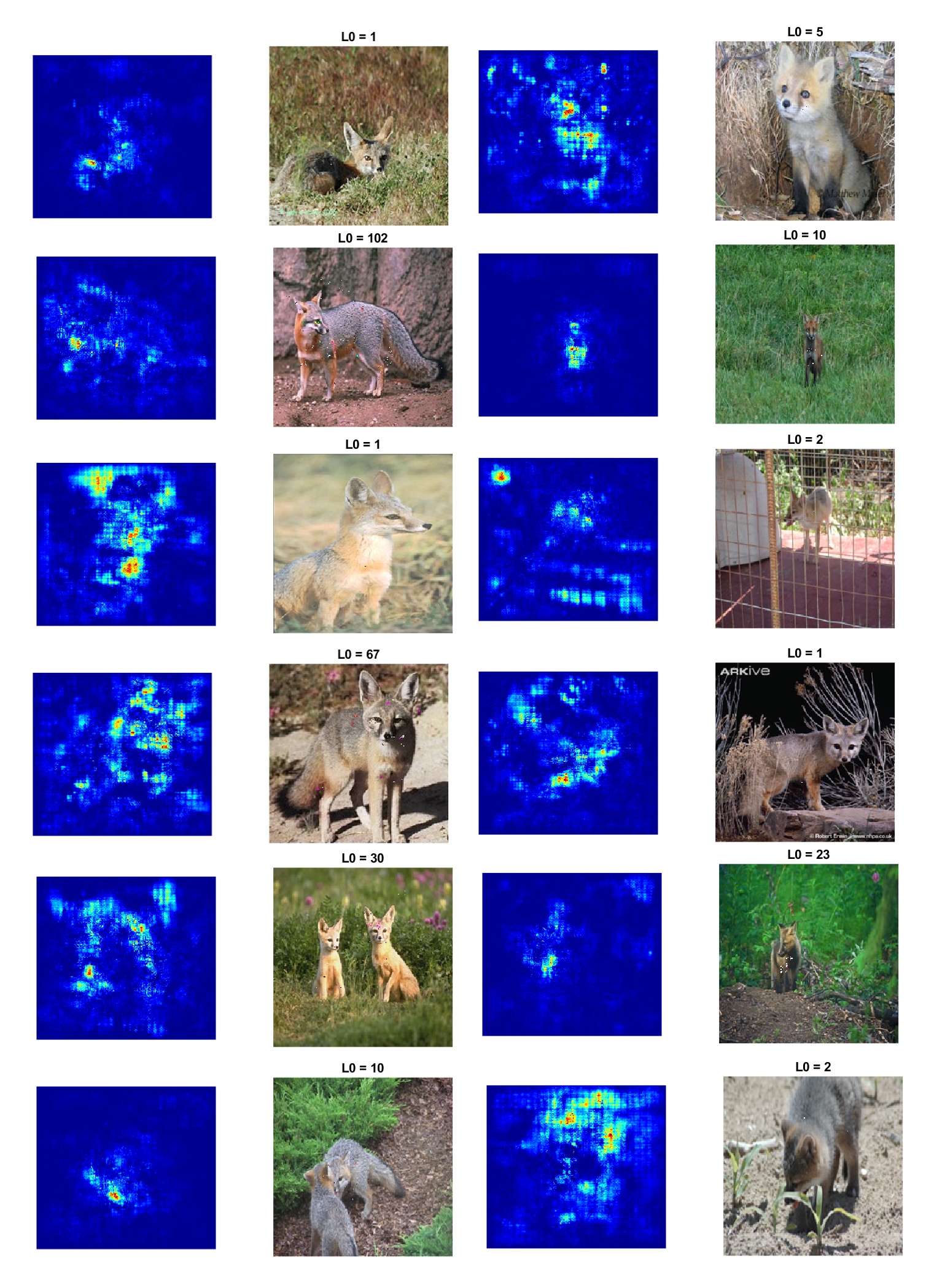}
	\caption{Adversarial examples on upper boundaries (right column) and their saliency maps (left column) for ImageNet AlexNet DNNs. Note that all adversarial images with $L_0 = 1, 2$ are also the ground-truth $L_0$-norm adversarial images since their upper bounds and lower bounds local robustness have converged.}
	\label{fig:casefour-4}
\end{figure}

\begin{figure}
	\includegraphics[width=1\linewidth]{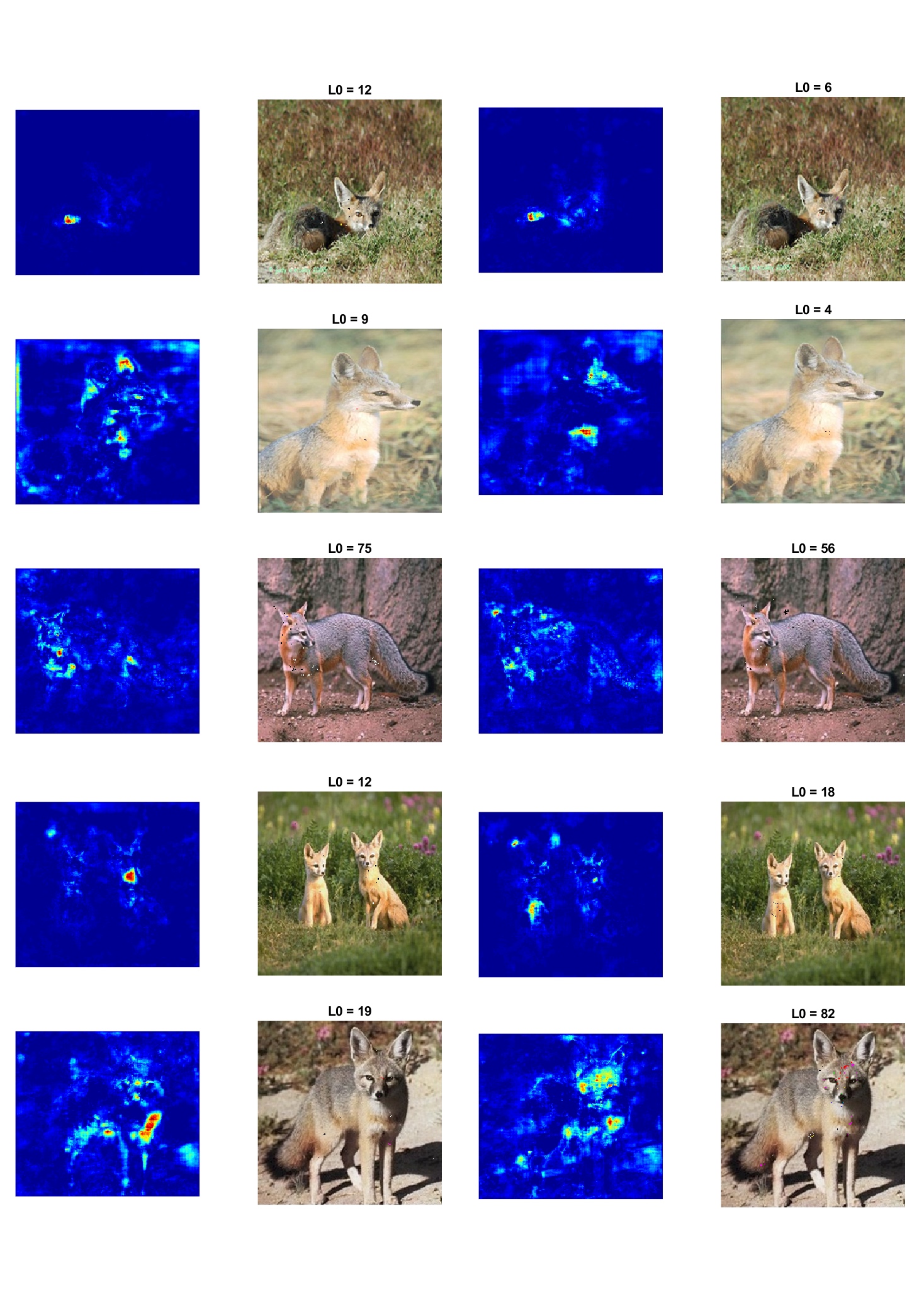}
	\caption{Adversarial examples on upper boundaries (right column) returned by {L0-TRE}, and their saliency maps (left column) for VGG-16 and VGG-19. The first and second columns are for VGG-16; the third and fourth columns are for VGG-19.}
	\label{fig:casefour-5}
\end{figure}

\end{document}